\documentclass[sigconf]{acmart}

\renewcommand\footnotetextcopyrightpermission[1]{} 

\AtBeginDocument{%
  \providecommand\BibTeX{{%
    \normalfont B\kern-0.5em{\scshape i\kern-0.25em b}\kern-0.8em\TeX}}}

\usepackage{subfigure}
\usepackage{bm}
\usepackage{color}
\usepackage{subfigure}
\usepackage{bm}
\usepackage{caption}
\usepackage{booktabs}
\usepackage{soul}
\usepackage{multirow}

\newcommand{\nop}[1]{}

\newtheorem{theorem}{Theorem}[section]

\usepackage{algorithmic}
\usepackage{graphicx}
\usepackage{textcomp}
\usepackage{xcolor}
\def\BibTeX{{\rm B\kern-.05em{\sc i\kern-.025em b}\kern-.08em
    T\kern-.1667em\lower.7ex\hbox{E}\kern-.125emX}}
    
\newcommand{\model}{\textsc{REASON}}

\newcommand\blfootnote[1]{%
  \begingroup
  \renewcommand\thefootnote{}\footnote{#1}%
  \addtocounter{footnote}{-1}%
  \endgroup
}

\begin{document}

\title{Hierarchical Graph Neural Networks for \\Causal Discovery and Root Cause Localization}

  \author{Dongjie Wang\textsuperscript{\rm 1,2,*,\dag}, Zhengzhang Chen\textsuperscript{\rm 1,*,\#}, Jingchao Ni\textsuperscript{\rm 3}, Liang Tong\textsuperscript{\rm 1},
  \\Zheng Wang\textsuperscript{\rm 4}, Yanjie Fu\textsuperscript{\rm 2}, Haifeng Chen\textsuperscript{\rm 1}} 
    \affiliation{ 
    \textsuperscript{\rm 1}\institution{NEC Laboratories America Inc.}
      \country{}
         \textsuperscript{\rm 2}\institution{University of Central Florida}
         \country{}
       \textsuperscript{\rm 3}\institution{Amazon Inc.}
      \country{ }
      \textsuperscript{\rm 4}\institution{University of Utah}
      \country{}
    }
\begin{abstract} 
The goal of root cause analysis is to identify the underlying causes of system problems by discovering and analyzing the causal structure from system monitoring data. 
It is indispensable for maintaining the stability and robustness of large-scale complex systems.
Existing methods mainly focus on the construction of a single effective isolated causal network, whereas many real-world systems are complex and exhibit interdependent structures (\textit{i.e.}, multiple networks of a system are interconnected by cross-network links).
In interdependent networks, the malfunctioning effects of problematic system entities can propagate to other networks or different levels of system entities. Consequently, ignoring the interdependency results in suboptimal root cause analysis outcomes.

In this paper, we propose \model, a novel framework that enables the automatic discovery of both intra-level (\textit{i.e.}, within-network) and inter-level (\textit{i.e.}, across-network) causal relationships for root cause localization. \model\ consists of Topological Causal Discovery and Individual Causal Discovery. The Topological Causal Discovery component aims to model the fault propagation in order to trace back to the root causes. 
To achieve this, we propose novel hierarchical graph neural networks to construct interdependent causal networks by modeling both intra-level and inter-level non-linear causal relations. Based on the learned interdependent causal networks, we then leverage random walk with restarts to model the network propagation of a system fault. The Individual Causal Discovery component focuses on capturing abrupt change patterns of a single system entity. 
This component examines the temporal patterns of each entity's metric data (\textit{i.e.}, time series), and estimates its likelihood of being a root cause based on the Extreme Value theory. Combining the topological and individual causal scores, the top $K$ system entities are identified as root causes. Extensive experiments on three real-world datasets with case studies demonstrate the effectiveness and superiority of the proposed framework.\blfootnote{$^*$The authors contribute equally.}\blfootnote{$^\dag$Work was done during an internship at NEC Laboratories America.}\blfootnote{$^\#$Corresponding author. Email: zchen@nec-labs.com.}
\end{abstract}

\maketitle
\pagestyle{empty}
\section{Introduction} 
Root Cause Analysis (RCA) refers to the process of identifying the root causes of system faults using surveillance metrics data~\cite{andersen2006root,kiciman2005root}.
It has been widely used in IT operations, industrial process control, telecommunications, {\em etc.}, because a failure or malfunction in these systems would drastically affect user experiences and result in financial losses. For instance, an intermittent outage of Amazon Web Services 
can result in 
a loss of around \$210 millions~\cite{amazon}.
To maintain the reliability and robustness of such systems, Key Performance Indicators (KPIs), such as latency or connection time in a microservice system, and metrics data, such as CPU/memory usages in a microservice system are often monitored and recorded 
in real-time for system diagnosis. 
The intricacy of these systems and the magnitude of the monitoring data, however, make manual root cause analysis unacceptably expensive and error-prone. 
Consequently, an efficient and effective root cause analysis that enables rapid service recovery and loss mitigation is essential for the steady operation and robust management of large-scale complex systems.

\begin{figure}[!t]
    \centering
    \includegraphics[width=0.92\linewidth]{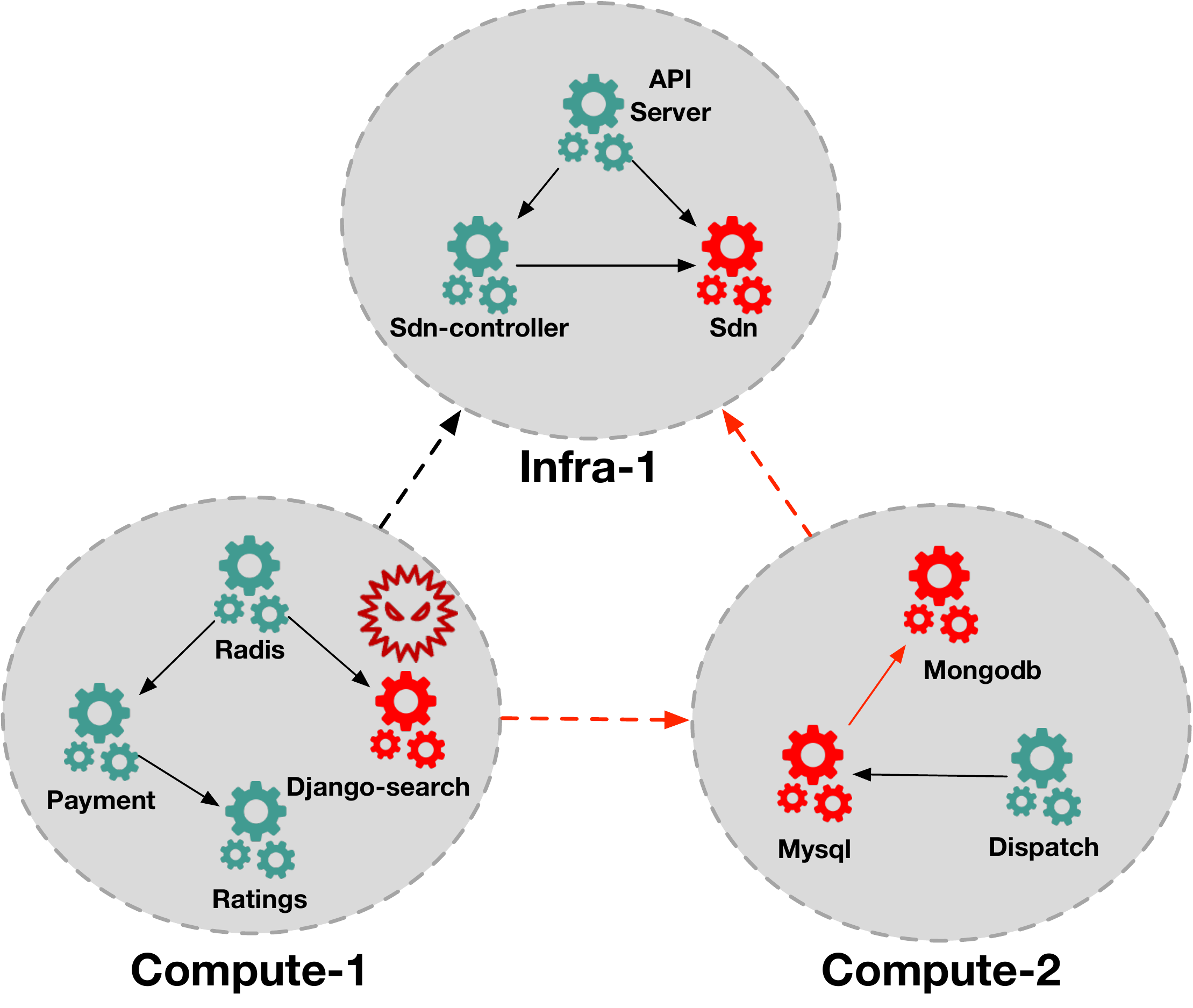}
    \captionsetup{justification=justified, singlelinecheck=off}
    \caption{ An example of interdependent causal networks. The main network is represented (\textit{i.e.}, server network) by dashed nodes and edges. The domain-specific networks (\textit{i.e.}, pod networks) are represented by solid nodes and edges. The edges highlighted by red colors indicate how the malfunctioning effects of a root cause propagate in the system. 
    }
    \label{fig:mcs}
\end{figure}


Prior studies on root cause analysis~\cite{liu2021microhecl,lin2018microscope,9521340,meng2020localizing} 
have mainly focused on a simplified scenario, where the target system is modeled as a single isolated causal graph, and the system's malfunctioning effects can only propagate within the same network of entities. For instance, to identify various types of service root causes, Liu \textit{et al.}~\cite{liu2021microhecl} generated a service call graph based on domain-specific software and rules. 
To discover the root causes of microservice system failures, Chen \textit{et al.}~\cite{lin2018microscope} constructed a directed acyclic graph that depicts the invoking relations among microservice applications. These methods have been applied to some uncomplicated systems with isolated network structures.    

However, a real-world complex system usually consists of multiple networks that coordinate 
in a highly complex 
manner~\cite{gao2014single,amini2020interdependent}. \nop{These networks interact with each other, 
and the failure of a system entity in one network could 
spread to the dependent entities in other networks, which in turn may result in 
cascading damages/failures~\cite{buldyrev2010catastrophic,liu2016breakdown} that could circulate the interconnected levels, 
with catastrophic consequences.} 
These networks are interconnected, and if one network's system entity fails, it may spread to its dependent entities in other networks, which may then cause cascading damages or failures~\cite{buldyrev2010catastrophic,liu2016breakdown} that could circulate throughout the interconnected levels with catastrophic consequences. For instance, Figure~\ref{fig:mcs} shows the malfunction of the pod \textit{Django-search} first spreads to the server network and causes the fault of the server \textit{Compute-1}; Then, the malfunctioning effects spread to the pod network of \textit{Compute-2} and causes the faults of the pod \textit{Mongodb} and \textit{Mysql}; Finally, the pod \textit{Sdn} in \textit{Infra-1} is also affected, resulting in the system failure. In this failure case, it is quite difficult to pinpoint 
the root cause \textit{Django-search} if we only model the server network (\textit{i.e.}, the three servers) or one of the three pod networks of the microservice system. 
\nop{In contrast, 
modeling the interconnected multi-network structures 
provides a more comprehensive understanding of the 
system 
and 
a more effective method for identifying the root causes. }
Thus, modeling the interconnected multi-network structures is vital for comprehensive understanding of the complex 
system and effective root cause localization.

\nop{\item \textbf{ \st{Issue 1: \st{requiring} excessive domain knowledge.} } 
{\color{red} Yanjie: every practical method needs domain knowledge. It is not a unique and articulate issue. maybe change your wording. }
For instance, Li \textit{et al.} located root cause via tracing on predefined invocation relations among different  microservices~\cite{9521340}. 
Liu \textit{et al.} identified different kinds of service anomalies using a service call graph generated by domain-specific software and rules~\cite{liu2021microhecl}.
These methods rely heavily on domain knowledge for detecting root causes. 
However, for many complex systems, it is expensive 
to acquire accurate domain knowledge, 
which makes these methods 
difficult to generalize to different domains. 
{\color{red} Yanjie: lacks a summary of this issue, dont just list a bunch of papers. }

\textbf{Issue 2: \st{ignoring} hierarchical structures.}
\nop{For example, Meng \textit{et al.} first discovered the causal relations among different time series, then utilized random walk to locate root cause metrics in a microservice~\cite{meng2020localizing}. 
Deng \textit{et al.} proposed an attention-based graph neural network to detect correlations among sensors and utilized attention scores to identify root causes in water treatment networks~\cite{deng2021graph}.
The structure learning part of these approaches is simple and ignores hierarchies in complex systems, which limits their effectiveness.}
Real-world complex systems usually
consist of different levels/layers of entities that coordinate 
in a highly complex 
manner. One example is the cloud computing facilities with microservice architecture (see Figure~\ref{fig:mcs}), which is typically constituted by 
hundreds of different levels of entities that vary from servers to application software. 
Previous researches~\cite{meng2020localizing,deng2021graph} 
mostly limit their focus to a simplified case, where the fault patterns 
only propagate within 
the same level of entities, 
ignoring the fact that the failure of a entity in one level could lead to the failure of the dependent entities in other levels, which in turn may result in 
cascading damages that could circulate the interconnected levels, 
with catastrophic consequences. For instance, Figure~\ref{fig:mcs} shows the malfunction of a pod causes the fault of a server machine, then the damaging effects spread to the whole system, resulting in the system failure. Thus, modeling the hierarchical structures 
is vital for a comprehensive understanding of a system 
and 
a more effective root cause analysis framework.
}
\nop{However, real-world complex systems (e.g. urban systems,  industrial systems, and microservice systems) are replete with hierarchical graph structures that are crucial for identifying root causes.
For instance, Figure ~\ref{fig:mcs} shows a root cause example in a microservice system, in which the malfunction of a pod causes the anomaly of a server machine, then the damaging effects spread to the whole system, resulting in the system failure.
But how to model such complex structural information in systems for root cause localization?}

\nop{Interdependent networks have been demonstrated to be an effective tool for modeling complex real-world systems across several disciplines, such as ranking, clustering, searching ~\cite{ni2014inside,}. 
In light of these impressive achievements, we adopt this graph data model to capture complex topological causal relations in real-world systems for localizing root causes.
To the best of our knowledge, we are the first to introduce this concept into the RCA domain.}

Recently, a promising approach 
for modeling such interconnected structures in complex systems has emerged through
the concept of interdependent networks (or network of networks)~\cite{liu2016breakdown,amini2019sustainable,NEKOVEE2007457,amini2020interdependent}. In interdependent networks, each node of the main network can be represented as a domain-specific network. Let us elaborate using the example in Figure~\ref{fig:mcs} again. Here, the dashed network represents a server/machine network (the main network), where the nodes are three different servers and edges/links indicate the causal relations among different servers. Each node of this main network is further represented as a pod network (the domain-specific network), where nodes are pods and edges denote their causal relations. Collectively, we call this structure a (server-pod) interdependent networks. And since all the edges in these interdependent networks indicate causal dependencies, we further call it \textit{interdependent causal networks}.\nop{the nodes \st{within}  one network are connected by \st{connectivity links} {\color{red} Yanjie: what are connectivity links?} (\textit{e.g.}, the friendships between individuals in a social network), and the nodes of different networks are adjacent to one another via \st{dependency links} {\color{red} Yanjie: you need to introduce this concept earlier, some descriptions are confusing. } (\textit{e.g.}(\textit{e.g.}, employer-employee relationships in a social network). }
Interdependent networks have been widely used in the study of various topics, including the academic influence of scholars~\cite{ni2014inside}, the spreading pattern of rumors in the complex social network~\cite{NEKOVEE2007457}, and etc. However, existing methods only consider physical or statistical correlations, but not causation, and thus cannot be directly applied 
for locating root causes.

Enlightened by the interdependent networks, this paper aims to learn interdependent causal relationships from monitoring metrics in multi-network systems for accurately identifying root causes when a system failure/fault occurs. Formally, given the system KPI data, the multi-level interconnected system entities, and their metrics data ({\em i.e.}, time series), our goal is to learn interdependent causal structures for discovering the root causes of system failures. There are two major challenges in this task:

\begin{itemize}
\item \textbf{Challenge 1: Learning interdependent causal networks and modeling fault propagation in interdependent causal networks.} As aforementioned, in real-world systems with interdependent networks structures, malfunctioning effects of root causes can propagate to other nodes of the same level or different levels (\textit{i.e.}, main network level and domain-specific network level), resulting in catastrophic failure of the entire system. To capture such propagation patterns for root cause localization, we need to learn the causal relationships not only within the same level but also across 
levels. After modeling the interdependent causal relationships, we still need to model the propagation of malfunctioning effects in the learned causal interdependent networks.  	
\item \textbf{Challenge 2: Identifying abrupt change patterns from the metrics data of an individual system entity.} In addition to the topological patterns, metrics data associated with the system entities can exhibit abrupt change patterns during the incidence of system faults, particularly those that are short-lived (\textit{e.g.}, fail-stop failures). The malfunctioning effects of the root cause may end quickly before they can spread. Thus, the temporal patterns from the metrics data can provide individual causal insights for locating root causes. The challenge is how to capture abrupt change patterns and determine the individual causal effect associated with the system failure. 

\end{itemize}

To address these 
challenges, in this paper, we propose \model, a generic inte\underline{r}depend\underline{e}nt c\underline{a}u\underline{s}al networks based framework, for root cause l\underline{o}calizatio\underline{n} in complex systems with interdependent network structures.
\model\ consists of Topological Causal Discovery (TCD) and Individual Causal Discovery (ICD). 
For the TCD component, the assumption is that the malfunctioning effects of root causes can propagate to other system entities of the same level or different levels over time~\cite{buldyrev2010catastrophic,liu2016breakdown}. To capture such propagation patterns, we propose a hierarchical graph neural networks based causal discovery method to discover both intra-level (\textit{i.e.}, within-network) and inter-level (\textit{i.e.}, across-network) non-linear causal relationships.
Then, we leverage a random walk with restarts to model the network
propagation of a system fault. 
The ICD component, on the other hand, focuses on individual causal effects, by analyzing the metrics data (\textit{i.e.}, time series) of each system entity. Especially considering the short-lived failure cases (\textit{e.g.}, fail-stop failures), there may be no propagation patterns.
We design an Extreme Value 
theory based method to capture the abrupt fluctuation patterns and estimate the likelihood of each entity being a root cause. 
Finally, we integrate the findings of TCD and ICD, and output the system entities with the top-$K$ greatest causal scores as the root causes.
Extensive experiments and case studies are conducted on three real-world datasets to validate the efficacy of our work.

\begin{figure*}[!t]
    \centering
    \includegraphics[width=0.95\linewidth]{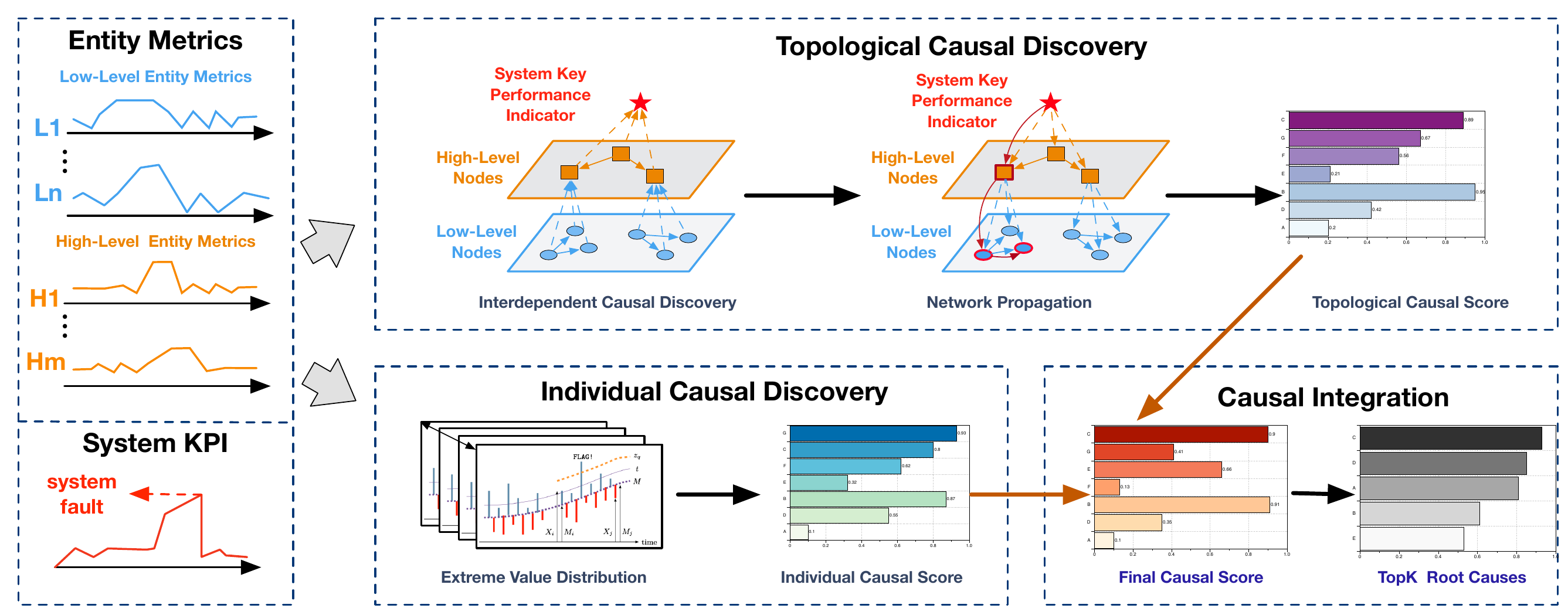}
    \captionsetup{justification=justified, singlelinecheck=off}
    \caption{The overview of the proposed framework \model. It consists of three major steps: topological causal discovery, individual causal discovery, and causal integration.\nop{This framework locates root causes based on monitoring entity metrics and system KPI.
    Individual causal discovery is to capture fluctuation patterns of root causes. 
    Topological causal discovery is to capture the propagation characteristics of root causes.
    Causal integration integrates the discovery results of the two branches for locating root causes accurately.}
    }
    \label{fig:framework}
\end{figure*}



\section{Preliminaries}
\label{sec:pre}
\nop{In this section, we present important definitions and the problem statement of our framework. }

\nop{\subsection{Important Definitions}}
\noindent \textbf{System Key Performance Indicator (KPI).}  A KPI is a monitoring time series that indicates the system status.
For instance, in a microservice system, the latency or connection time can be used to assess the system status. The smaller the latency, the higher the system's performance quality. If the connection time is too long, it is likely that the system has failed.

\noindent \textbf{Entity Metrics.} Entity metrics data can be collected by monitoring different levels of system entities. It usually contains a number of metrics, which indicate the status of a system's underlying entity. For example, in a microservice system, the underlying entity can be a physical machine, container, virtual machine, pod, etc. And the corresponding metrics can be CPU utilization, memory utilization, disk IO utilization, etc. The data for all these metrics are essentially time series. An anomalous metric of a microservice’s underlying entity can be the potential root cause of an anomalous system latency/connection time, which indicates a microservice failure.

\noindent \textbf{Interdependent Networks (INs)}. 
Interdependent networks model the interconnections of multiplex networks \cite{ni2014inside,10.1145/2783258.2783262}. Given a $g \times g$ main network $\mathbf{G}$, a set of domain-specific networks $\mathcal{A} = \{\mathbf{A}_1, \cdots, \mathbf{A}_g \}$, and an edge set $\mathbf{\ddot{E}}$ that represents the edges between the nodes in $\mathcal{A}$ and the nodes in $\mathbf{G}$,  
INs are 
defined as a triplet $\mathcal{R}=<\mathbf{G},\mathcal{A}, \mathbf{\ddot{E}}>$.
\nop{In addition, we represent the node set in $\mathbf{G}$ as $\mathbf{V}^G$ and 
the node set in $\mathcal{A}$ as $\mathbf{V}^{\mathcal{A}} = (\mathbf{V}^{A_1}, \cdots, \mathbf{V}^{A_g})$.
In this paper, we regard high-level entities as $\mathbf{V}^G$ and low-level entities as $\mathbf{V}^A$.}
The node set in $\mathbf{G}$, {\em a.k.a.} high-level nodes, is denoted by $\mathbf{V}^G$, and
the node set in $\mathcal{A}$, {\em a.k.a.} low-level nodes, is denoted by $\mathbf{V}^{\mathcal{A}} = (\mathbf{V}^{A_1}, \cdots, \mathbf{V}^{A_g})$. As a special type of INs, interdependent causal networks represent the INs with the edges indicating causal relations. 

Take Figure~\ref{fig:mcs} as an example, the dashed network is the main network $\mathbf{G}$, which has three server nodes, including \textit{Compute-1}, \textit{Compute-2}, and \textit{Infra-1}.
Each of these main nodes contains a domain-specific network that is made up of several applications/pods.
For instance, the main node \textit{Compute-2} is further represented as a domain-specific network with three pod nodes, including \textit{Mysql}, \textit{Mongodb}, and \textit{Dispatch}. And the solid edges indicate the causal relationships between different pods, while the dashed edges indicate the causal relationships between different servers.

Without loss of generality, we focus on
two levels of system entities.
Our goal is to identify root causes by 
automatically learning interdependent causal relations between different levels of system entities and the system KPI.
The identified root causes are low-level system entities to reflect fine-grained root cause detection.

\noindent \textbf{Problem Statement}. 
Given metrics/sensor data of multi-level system entities corresponding to high-level and low-level nodes in main and domain-specific networks $\{\mathbf{X}^G, \mathbf{X}^{\mathcal{A}}\}$, and system key performance indicator $\mathbf{y}$, the problem is to construct an interdependent causal network $\mathcal{R}=<\mathbf{G},\mathcal{A}, \mathbf{\ddot{E}}>$, and identify the top $K$ low-level nodes in $\mathbf{V}^\mathcal{A}$ that are most relevant to $\mathbf{y}$.

\nop{
\textbf{Given:} 
(1) Measurement time series of both nodes (system entities) in main and domain-specific networks  $\{\mathbf{X}^G, \mathbf{X}^{\mathcal{A}}\}$;
(2) System status indicator $\mathbf{y}$;

\textbf{Find:}
(1) Causal interdependent networks  $\mathcal{R}=<\mathbf{G},\mathcal{A}, \theta>$;
(2) Top K nodes in $\mathcal{V}^\mathcal{A}$ that are most relevant to $\mathbf{y}$.

}

\section{Methodology}
We present \model, an interdependent causal network based framework for root cause localization. As illustrated in Figure~\ref{fig:framework}, \model\ includes three major steps: 1) topological causal discovery; 2) individual causal discovery; and 3) causal integration.

\subsection{Topological Causal Discovery}
\label{topological_rca}
Root causes ({\em i.e.,} the system entities that cause the system failures or faults) could propagate malfunctioning or fault effects to other system entities of the same network or across different networks over time~\cite{buldyrev2010catastrophic,ash2007optimizing,liu2016breakdown}, which makes real root causes hard to locate.
To address this challenge, we propose a hierarchical graph neural network based causal discovery method to construct interdependent causal graphs  among low-level and high-level system entities.
Failure propagation is modeled on the learned causal structures to provide topological guidance for locating root causes by simulating the malfunctioning effects of root causes.

\subsubsection{Hierarchical Graph Neural Network based Interdependent Causal Discovery}
\label{subsec:icd}
There can be more than one entity metric (\textit{i.e.}, multi-variate time series) per system entity (refer to Section~\ref{sec:pre}).
For each individual metric, we learn interdependent causal graphs among different system entities using the same learning strategy.
To ease the description, we take one metric of system entities as an example to illustrate the interdependent causal discovery process.

The metric of system entities (\textit{i.e.}, high-level or low-level) is a multivariate time series $\{\mathbf{x}_0, \cdots, \mathbf{x}_T\}$.
The metric value at the $t$-th time step is $\mathbf{x}_t \in \mathbb{R}^{d}$, where $d$ is the  number of entities. 
The data can be modeled using the VAR model~\cite{tank2021neural,stock2001vector}, whose formulation is given by: 
\begin{equation}
    \mathbf{x}^\top_t =  \mathbf{x}^\top_{t-1}\mathbf{B}_1 + \cdots + \mathbf{x}^\top_{t-p}\mathbf{B}_p + \bm{\epsilon}^\top_t, \quad  t=\{p,\cdots, T\}\label{svar1}
\end{equation}
where $p$ is the time-lagged order, $\bm{\epsilon}_t$ is the vector of error variables that are expected to be non-Gaussian and independent in the temporal dimension, $\{\mathbf{B}_1,\cdots, \mathbf{B}_p \}$ are the weighted matrix of time-lagged data. 
In the VAR model, the time series at $t$, $\mathbf{x}_t$, is assumed to be a linear combination of the past $p$ lags of the series.

Assuming that  $\{\mathbf{B}_1,\cdots, \mathbf{B}_p \}$ is constant across time, the Equation ~\eqref{svar1} can be extended into a matrix form:
\begin{equation}
    \mathbf{X} = \mathbf{\Tilde{X}}_1\mathbf{B}_1 + \cdots + \mathbf{\Tilde{X}}_p\mathbf{B}_p + \bm{\varepsilon}\label{svar2}
\end{equation}
where $\mathbf{X} \in \mathbb{R}^{m\times d}$ is a matrix and its each row is $\mathbf{x}^\top_t$;
$\mathbf{\{\Tilde{X}}_1,\cdots,\mathbf{\Tilde{X}}_p\}$ are the time-lagged data.

To simplify Equation~\ref{svar2}, let $\mathbf{\Tilde{X}} = [ \mathbf{\Tilde{X}}_1| \cdots | \mathbf{\Tilde{X}}_p]$ with its shape of $\mathbb{R}^{m\times pd}$ and $\mathbf{B} = [\mathbf{B}_1 | \cdots | \mathbf{B}_p]$ with its shape of $\mathbb{R}^{m\times pd}$.
Here, $m=T+1-p$ is the effective sample size, because the first $p$ elements in the metric data have no sufficient time-lagged data to fit  Equation~\ref{svar2}.
After that, we apply the QR decomposition to the weight matrix $\mathbf{B}$ to transform Equation~\ref{svar2} as follows:
\begin{equation}
    \mathbf{X} =  \mathbf{\Tilde{X}}\mathbf{\hat{B}}\mathbf{W} + \bm{\varepsilon}\label{svar3}
\end{equation}
where $\mathbf{\hat{B}} \in \mathbb{R}^{m\times pd}$ is the weight matrix of time-lagged data in the temporal dimension; $\mathbf{W} \in \mathbb{R}^{d \times d}$ is the weighted adjacency matrix, which reflects the relations among system entities.

A nonlinear autoregressive model allows $\mathbf{x}_t$ to evolve according to more general nonlinear dynamics~\cite{billings2013nonlinear}. In a forecasting setting, one promising way is to jointly model the nonlinear functions using neural networks~\cite{billings2013nonlinear,li2017diffusion}. By applying neural networks $f$ to Equation~\ref{svar3}, we have 
\begin{equation}
    \mathbf{X} = f(\mathbf{\Tilde{X}}\mathbf{\hat{B}}\mathbf{W};\bm{\Theta}) + \bm{\varepsilon}
\end{equation}
where $\bm{\Theta}$ is the set of parameters of $f$.

Given the data $\mathbf{X}$ and $\mathbf{\Tilde{X}}$, here our goal is to estimate weighted adjacency matrices $\mathbf{W}$ that correspond to directed acyclic graphs (DAGs). The causal edges in $\mathbf{W}$ go only forward in time, and thus they do not create cycles. In order to ensure that the whole network is acyclic, it thus suffices to require that $\mathbf{W}$ is acyclic. Minimizing the least-squares loss with the acyclicity constraint gives the following optimization problem:

\begin{equation}
    \text{min} \, \frac{1}{m} \left \| \mathbf{X} - f(\mathbf{\Tilde{X}}\mathbf{\hat{B}}\mathbf{W};\bm{\Theta})  \right \|^2 \quad s.t. \ \mathbf{W} \text{ is acyclic}\label{loss1},
\end{equation}

To learn $\mathbf{W}$ in an adaptive manner, we adopt the following layer:
\begin{equation}
    \mathbf{W} = \text{RELU}(\text{tanh}(\mathbf{W}_+\mathbf{W}_-^\top - \mathbf{W}_-\mathbf{W}_+^\top)),
\end{equation}
where $\mathbf{W}_+ \in \mathbb{R}^{d\times d}$ and $\mathbf{W}_- \in \mathbb{R}^{d\times d}$ are two parameter matrices. 
This learning layer aims to enforce the asymmetry of $\mathbf{W}$, because 
the propagation of malfunctioning effects is unidirectional and acyclic from root causes to subsequent entities.
In the following sections, $\mathbf{W}^G$ denotes the causal relations between high-level nodes and $\mathbf{W}^{\mathcal{A}}$ denotes the causal relations between low-level nodes.

\begin{figure}[!t]
    \centering
    \includegraphics[width=0.95\linewidth]{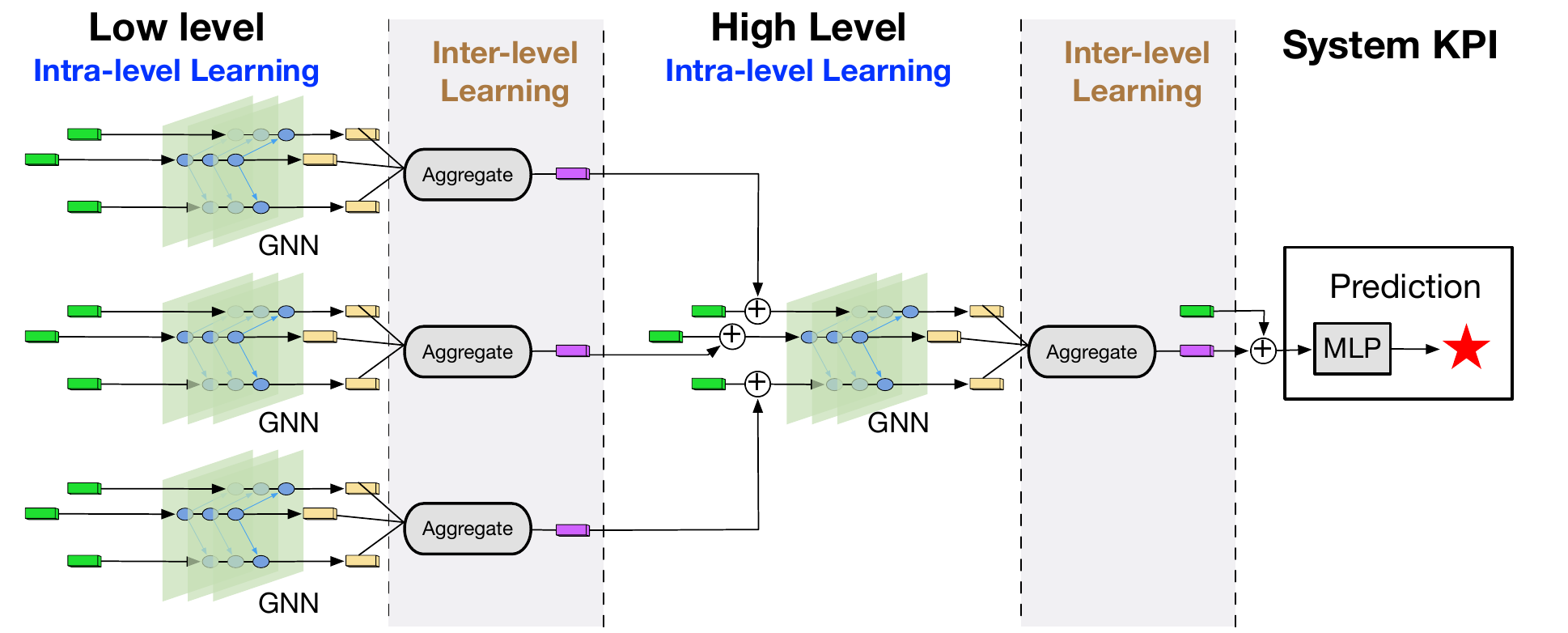}
    \captionsetup{justification=justified, singlelinecheck=off}
    \caption{The learning process of hierarchical GNNs. Intra-level learning captures causation within the same-level system entities. Inter-level learning aggregates low-level information to high-level for constructing cross-level causation.}
   
    \label{fig:hcgl}
\end{figure}

Then, the causal structure learning  for the interdependent networks can be divided into intra-level learning and inter-level learning. 
Intra-level learning is to learn the causation among the same level of nodes, while inter-level learning is to learn the cross-level causation. To model the influence of low-level nodes on high-level nodes, we aggregate low-level information into high-level nodes in inter-level learning.  
Figure~\ref{fig:hcgl} shows the  learning process. 

For \textbf{intra-level learning}, we adopt the same learning strategy to learn causal relations among both high-level nodes and low-level nodes.
Specifically, we first apply $L$ layers of GNN to the time-lagged data $\{\mathbf{x}_{t-1},\cdots,\mathbf{x}_{t-p}  \}\in \mathbb{R}^{d\times p}$ to obtain its embedding.
In the $l$-th layer, the embedding $\mathbf{z}^{(l)}$ is obtained by aggregating the nodes' embedding and their neighbors' information at the $l-1$ layer.
Then, the embedding at the last layer $\mathbf{z}^{(L)}$ is used to predict the metric value at the time step $t$ by an MLP layer.
This process can be represented as
\begin{equation}
\left\{
             \begin{array}{lr}
             \mathbf{z}^{(0)} = [\mathbf{x}_{t-1},\cdots, \mathbf{x}_{t-p}], \\
             \mathbf{z}^{(l)} = \text{GNN}(\text{Cat}(\mathbf{z}^{(l-1)}, \mathbf{W}\cdot \mathbf{z}^{(l-1)} )\cdot\mathbf{B}^{(l)}),
              \\
              \breve{\mathbf{x}}_t = \text{MLP}(\mathbf{z}^{(L)};\bm{\Theta}),
             \end{array}
\right.
\label{gnn_emb}
\end{equation}
where Cat is the concatenation operation; $\mathbf{B}^{(l)}$ is the weight matrix of the $l$-th layer;
GNN is activated by the RELU function to capture non-linear correlations in the time-lagged data. 
Our goal is to minimize the difference between the actual value $\mathbf{x}_t$ and the predicted value $\breve{\mathbf{x}}_t$.
Thus, the optimization objective is defined as follows 
\begin{equation}
    \mathcal{L} = \frac{1}{m }\sum\limits_t( \mathbf{x}_t - \breve{\mathbf{x}}_t )^2 \label{loss2}
\end{equation}

As shown in Figure~\ref{fig:hcgl}, we conduct  intra-level learning for the low-level and high-level system entities for constructing $\mathbf{W}^{\mathcal{A}}$ and $\mathbf{W}^G$, respectively.
The optimization objectives for the low-level and high-level causal relations, in the same format as  Equation~\ref{loss2}, are denoted by $\mathcal{L}_{\mathcal{A}}$ and $\mathcal{L}_{G}$, respectively.

For \textbf{inter-level learning}, we aggregate the information of low-level nodes to the high-level nodes for constructing the cross-level causation.
So, the initial embedding of high-level nodes $\mathbf{\ddot{z}}^{(0)}$ is the concatenation of their time-lagged data $\{\mathbf{\ddot{x}}_{t-1},\cdots,\mathbf{\ddot{x}}_{t-p} \}$ and aggregated low-level embeddings, which can be formulated as follows
\begin{equation}
    \mathbf{\ddot{z}}^{(0)} = \text{Cat}([\mathbf{\ddot{x}}_{t-1},\cdots,\mathbf{\ddot{x}}_{t-p}],\mathbf{\ddot{W}}\cdot\mathbf{z}^{(L)})
\end{equation}
where $\mathbf{\ddot{W}}$ is a weight matrix that controls the contributions of low-level embeddings to high-level embeddings.
As shown in Figure~\ref{fig:hcgl}, there are two inter-level learning parts.
The first one is used to learn the cross-level causal relations between  low-level  and high-level nodes, denoted by $\mathbf{\ddot{W}}^{\mathcal{A}G}$.
The second one is used to construct the causal linkages between high-level nodes and the system KPI, denoted by $\mathbf{\ddot{W}}^{GS}$.
During this process, we predict the value of the system KPI at the time step $t$ and aim to make the predicted values close to the actual ones.
Hence, we formulate the optimization objective $\mathcal{L}_{S}$, whose format is the same as Equation~\ref{loss2}.

In addition, the learned interdependent causal graphs must 
meet the acyclicity requirement.  Since the cross-level causal relations $\mathbf{\ddot{W}}^{\mathcal{A}G}$ and $\mathbf{\ddot{W}}^{GS}$ are unidirectional, 
only $\mathbf{W}^{\mathcal{A}}$ and $\mathbf{W}^G$ need to be acyclic.
To achieve this goal, inspired by the work~\cite{zheng2018dags}, we use the trace exponential function: $h(\mathbf{W}) = tr(e^{\mathbf{W}\circ \mathbf{W}})-d=0$ that satisfies $h(\mathbf{W})=0$ if and only if $\mathbf{W}$ is acyclic.
Here, $\circ$ is the Hadamard product of two matrices. Meanwhile, to enforce the sparsity of $\mathbf{W}^{\mathcal{A}}$, $\mathbf{W}^G$, $\mathbf{\ddot{W}}^{\mathcal{A}G}$, and $\mathbf{\ddot{W}}^{GS}$ for producing robust causation, we use the $L1$-norm to regularize them.
So, the final optimization objective is
\begin{equation}
\begin{aligned}
    & \mathcal{L}_{final} = \,  (\mathcal{L}_{\mathcal{A}} +
    \mathcal{L}_{G} + \mathcal{L}_{S})
    \\& + 
    \lambda_1 (\left \| \mathbf{W}^{\mathcal{A}} 
    \right \|_1 + \left \| \mathbf{W}^{G} 
    \right \|_1 + \left \| \mathbf{\ddot{W}}^{\mathcal{A}G} 
    \right \|_1 + \left \| \mathbf{\ddot{W}}^{GS} 
    \right \|_1)
    \\&
    + \lambda_2 (h(\mathbf{W}^{\mathcal{A}}) + h(\mathbf{W}^{G}))
\end{aligned}
\end{equation}
where $\left \| \cdot \right \|_1$ is the element-wise $L1$-norm; $\lambda_1$ and $\lambda_2$ are two parameters that control the contribution of regularization items.
We aim to minimize $\mathcal{L}_{final}$ through the L-BFGS-B solver.
When the model converges, we construct interdependent causal networks 
through $\mathbf{W}^{\mathcal{A}}$, $\mathbf{W}^G$, $\mathbf{\ddot{W}}^{\mathcal{A}G}$, and $\mathbf{\ddot{W}}^{GS}$.

\subsubsection{Network Propagation on Interdependent Causal Graphs}
\label{sec_network_prop}
As aforementioned, starting from the root cause entity, malfunctioning effects will propagate to neighboring entities~\cite{cheng2016ranking}, and different types of system faults can trigger diverse propagation patterns. 
This observation motivates us to apply network propagation to the learned causal structure to mine the hidden actual root causes. 

The learned interdependent causal structure is a directed acyclic graph, which reflects the causal relations from the low-level to the high-level to the system level.
In order to trace back the root causes, we need to conduct a reverse analysis process.
Thus, we transpose the learned causal structure to get $<<\mathbf{G}^{\top}, \mathcal{A}^{\top}, \mathbf{\ddot{E}}>, \text{KPI}>$\footnote{Here, $\mathbf{\ddot{E}}$ contains not only the edges between the nodes in $\mathcal{A}$ and the nodes in $\mathbf{G}$ but also the edges between the nodes in $\mathbf{G}$ and the node of system KPI.}, then apply a random walk with restart on the interdependent causal networks to estimate the topological causal score of each entity.

Specifically, the transition probabilities of a particle on the transposed structure can be denoted by 
\begin{equation}
    H = \begin{bmatrix}
  H_{\mathbf{G}\mathbf{G} } & H_{\mathbf{G}\mathcal{A}}\\
  H_{\mathcal{A}\mathbf{G}} & H_{\mathcal{A}\mathcal{A}}
\end{bmatrix}
\end{equation}
where $H_{\mathbf{G}\mathbf{G}}$ and $H_{\mathcal{A}\mathcal{A}}$ depict the walks within the same-level network. 
$H_{\mathbf{G}\mathcal{A}}$ and $H_{\mathcal{A}\mathbf{G}}$ describe the walks across different level networks.
Imagine that from the KPI node, a particle begins to visit the networks.
The particle randomly selects a high-level or low-level node to visit, then the particle either jumps to the low-level nodes or walks in the current graph with a probability value $\Phi \in [0,1]$.
The higher the value of $\Phi$ is, the more possible the jumping behavior occurs.
In detail, if a particle is located at a high-level node $i$ in $\mathbf{G}$, the probability of the particle moving to the high-level node $j$ is 
\begin{equation}
    H_{\mathbf{GG}}(i,j) = (1-\Phi)\mathbf{G}^{\top}(i,j)/\sum_{k=1}^g \mathbf{G}^{\top}(i,k)
\end{equation}
or jumping to the low-level node $b$ with a probability
\begin{equation}
    H_{\mathbf{G}\mathcal{A}}(i,b) = \Phi \mathbf{\ddot{W}}(i,b)/ \sum_{k=1}^{gd} \mathbf{\ddot{W}}(i,k)
\end{equation}
We apply the same strategy when the particle is located at a low-level node.
The particle walking between different low-level nodes has a visiting probability of $H_\mathcal{AA}$, whose calculation equation is similar to $H_{\mathbf{GG}}$.
Moreover, the visiting probability from a low-level node to a high-level node is $H_{\mathcal{A}\mathbf{G}}$, whose calculation equation is similar to $H_{\mathbf{G}\mathcal{A}}$.
The probability transition evolving equation of the random walk with restart can be formulated as
\begin{equation}
    \mathbf{\Tilde{p}}^{\top}_{t+1} = (1-\varphi)\mathbf{\Tilde{p}}_t^{\top} + \varphi\mathbf{\Tilde{p}}^{\top}_{rs}
\end{equation}
where $\mathbf{\Tilde{p}}^{\top}_{t+1} \in \mathbb{R}^{g+gd}$ and  $\mathbf{\Tilde{p}}_t^{\top}\in \mathbb{R}^{g+gd}$ are the visiting probability distribution at different time steps;
$\mathbf{\Tilde{p}}^{\top}_{rs} \in \mathbb{R}^{g+gd}$ is the initial visiting probability  distribution that depicts the visiting possibility of high-level or low-level nodes at the initialization step.
$\varphi \in [0,1]$ is the restart probability.
When the visiting probability distribution is convergence, we regard the probability score of the low-level nodes as the associated topological causal score.


\subsection{Individual Causal Discovery}
In addition to the topological causal effects, the entity metrics of root causes themselves could fluctuate stronger than those of other system entities during the incidence of some system faults. And for some short-lived failure cases (\textit{e.g.}, fail-stop failure), there may even be no propagation patterns.
Thus, we propose to individually analyze such temporal patterns in order to  provide individual causal guidance for locating root causes.

Compared with the values of entity metrics in normal time, the fluctuating values are extreme and infrequent.
Inspired by~\cite{siffer2017anomaly}, such extreme value follows the extreme value distribution, which is defined as:
\begin{equation}
    U_{\zeta}: x \rightarrow exp(-(1+\zeta x)^{-\frac{1}{\zeta}}), \quad \zeta \in \mathbb{R}, \quad 1+\zeta x > 0.
\end{equation}
where $x$ is the original value and  $\zeta$ is the extreme value index depending on the distribution of $x$.
Let the probability of potential extreme value in $x$ 
be $q$, the boundary\footnote{The boundary can be upper bound or lower bound of normal values.} $\varrho$ of normal value 
can be calculated through $\mathbb{P}(X>\varrho) = q$ based on $U_{\zeta}$.
However, since the distribution of $x$ is unknown, 
$\zeta$ should be estimated. 
The
Pickands-Belkema-de Haan theorem~\cite{pickands1975statistical} provides 
an approach to estimate $\zeta$, which is defined as follows:
\begin{theorem}
\textit{
The extrema of a cumulative distribution F converge to the distribution of $U_{\zeta}$, denoted as $F\in D_{\zeta}$,
if and only if a function $\delta$ exists, for all $x \in \mathbb{R}$ s.t. $1+\zeta x > 0$}:
\begin{equation}
    \frac{\overline{F}(\eta+\delta(\eta)x)}{\overline{F}(\eta)} \quad
    \overrightarrow{\eta\rightarrow\tau} 
    \quad
    (1+\zeta x)^{-\frac{1}{\zeta}},
\end{equation}
\end{theorem} 
where $U_{\zeta}$ refers to the extreme value distribution; $D_{\zeta}$ refers to a Generalized Pareto Distribution; $\eta$ is a threshold for peak normal value; and $\tau$ is the boundary of the initial distribution.
Assuming that $\eta$ is a threshold for peak normal value,  $X - \eta$ follows a Generalized Pareto Distribution (GPD) with parameters $\zeta$ and $\delta$ according to the theorem, which is defined as:  
\begin{equation}
   \mathbb{P}(X-\eta > x | X > \eta) \sim (1+\frac{\zeta x}{\delta(\eta)})^{-\frac{1}{\zeta}}.
\end{equation}
We can utilize the maximum likelihood estimation method~\cite{beirlant2004statistics} to estimate $\zeta$ and $\delta$.
Then, the boundary value $\varrho$ can be calculated by 
\begin{equation}
    \varrho \simeq \eta + \frac{\delta}{\zeta} ((\frac{qn}{N_{\eta}})^{-\zeta}-1).
    \label{eq:boundary}
\end{equation}
where $\eta, q$ can be provided by domain knowledge, $n$ is the total number of observations, and $N_{\eta}$ is the number of peak values ({\em i.e.}, the number of $X >\eta$).

Individual causal discovery is devised based on Equation~\eqref{eq:boundary}. Specifically, we divide the metric data of one system entity into two segments.
The first segment is used for initialization, and the second one is used for detection.
For initialization, we first provide the probability of the extreme value $q$ and the threshold of the peak value $\eta$ using a mean excess plot-based method~\cite{beirlant2004statistics}.
Then, we use the first time segment to estimate the boundary $\varrho$ of normal value according to Equation~\eqref{eq:boundary}.
Here, $\eta$ should be lower than $\varrho$.
For detection, we compare each value in the second time segment with $\varrho$ and $\eta$.
If the value is larger than $\varrho$, the value is abnormal, so we store it.
If the value is less than $\varrho$ but larger than $\eta$, which means the boundary $\varrho$ has been changed.
Hence, we add it to the first segment and re-evaluate the parameters $\zeta$ and $\delta$ to get new boundaries.
If the value is less than $\eta$, it is normal, so we 
ignore it.
Finally, we can collect all abnormal values and normalize them using the Sigmoid function.
The mean of the normalized values is regarded as the individual causal score of the associated system entity.


\subsection{Causal Integration}
Finally, we integrate the individual and topological causal scores of low-level system entities through the integration parameter $0\leq \gamma \leq 1$, which can be represented as $\mathbf{q}_{final} = \gamma\mathbf{q}_{indiv} + (1-\gamma)\mathbf{q}_{topol}$.
After that, we rank low-level nodes using $\mathbf{q}_{final}$ and select the top $K$ results as the final root causes.




\section{Experiments}

\nop{We now present the experimental results on three real-world datasets in order to validate the effectiveness of our framework (\model) for root cause localization.

We aim to answer the following questions:
\begin{itemize}
    \item \textbf{Q1} Can our framework (\model) effectively localize the root causes of system events?
    \item \textbf{Q2} How do the hierarchies that exist in real-world systems influence identifying root causes?
    \item \textbf{Q3}  Is the network propagation step indispensable for accurately finding root causes?
    \item \textbf{Q4} Is it significant to combine the results of individual and topological causal discovery for locating root causes?
    \item \textbf{Q5} How does the proposed interdependent causal discovery method (GNN) affect localizing root causes?
    \item \textbf{Q6}  How do different hyperparameter settings influence the performance of root cause localization?
    \item \textbf{Q7} What does the learned causal structure look like?
\end{itemize}
}

\begin{table*}[!ht]
\renewcommand\arraystretch{1}
	\centering
	\caption{Overall performance {\it w.r.t.} Swat dataset.} 
	\begin{tabular}{c|c|c|c|c|c|c|c|c|c|c}
		\hline
		 & PR@1 & PR@3 & PR@5    & PR@7 & PR@10   & MAP@3      & MAP@5 & MAP@7     & MAP@10 & MRR        \\ \hline
\model & \textbf{25.0\%} &  \textbf{28.13\%}   & \textbf{66.67\%} & \textbf{76.04\%} & \textbf{84.38\%} & \textbf{23.96\%}  & \textbf{35.0\%} & \textbf{46.73\%} & \textbf{57.60\%} &     \textbf{40.99\%}        \\
GNN & 18.75\% &  19.79\%   & 43.75\% & 52.08\% & 62.50\% & 18.06\%  & 27.92\% & 33.63\% & 41.88\% &     34.77\%        \\
PC & 12.5\% &  13.54\%   & 34.38\% & 47.92\% & 58.33\% & 12.85\%  & 20.42\% & 26.64\% & 35.0\% &     26.16\%        \\
C-LSTM & 12.5\% &  13.54\%   & 28.13\% & 40.63\% & 52.08\% & 13.89\%  & 17.71\% & 23.81\% & 31.88\% &     29.35\%        \\
Dynotears & 12.5\% &  29.17\%   & 32.29\% & 34.38\% & 42.71\% & 20.14\%  & 24.38\% & 26.93\% & 30.83\% &     27.85\%        \\
GOLEM & 6.25\% &  7.29\%   & 12.5\% & 39.58\% & 47.92\% & 7.64\%  & 9.58\% & 16.96\% & 25.0\% &     22.36\%        \\
\hline
	\end{tabular}
	\label{tab:overall_swat}
\end{table*}

\begin{table*}[!ht]
\renewcommand\arraystretch{1}
	\centering
	\caption{Overall performance {\it w.r.t.} WADI dataset.} 
	\begin{tabular}{c|c|c|c|c|c|c|c|c|c|c}
		\hline
		 & PR@1 & PR@3 & PR@5    & PR@7 & PR@10   & MAP@3      & MAP@5 & MAP@7     & MAP@10 & MRR        \\ \hline
\model & \textbf{28.57\%} &  \textbf{59.52\%}   & \textbf{65.0\%} & \textbf{76.19\%} & \textbf{79.76\%} & \textbf{42.46\%}  & \textbf{50.62\%} & \textbf{57.41\%} & \textbf{63.76\%} &     \textbf{53.35\%}        \\
GNN & 14.28\% &  26.19\%   & 34.28\% & 42.86\% & 54.76\% & 21.83\%  & 25.31\% & 30.15\% & 37.54\% &     32.71\%        \\
PC & 7.14\% &  27.38\%   & 35.0\% & 44.05\% & 50.0\% & 16.27\%  & 23.90\% & 28.47\% & 34.57\% &     27.74\%        \\
C-LSTM & 0\% &  20.24\%   & 35.0\% & 47.62\% & 51.19\% & 11.51\%  & 18.55\% & 25.83\% & 32.73\% &     24.40\%        \\
Dynotears & 7.14\% &  14.29\%   & 30.00\% & 29.76\% & 47.62\% & 10.71\%  & 17.43\% & 20.95\% & 26.81\% &     22.23\%        \\
GOLEM & 0\% &  19.05\%   & 40.0\% & 46.43\% & 53.57\% & 9.92\%  & 20.38\% & 27.82\% & 34.83\% &     23.48\%        \\
\hline
	\end{tabular}
	\label{tab:overall_wadi}
\end{table*}

\begin{table*}[!ht]
\renewcommand\arraystretch{1}
	\centering
	\caption{Overall performance {\it w.r.t.} AIOps dataset.} 
	\begin{tabular}{c|c|c|c|c|c|c|c|c|c|c}
		\hline
		 & PR@1 & PR@3 & PR@5    & PR@7 & PR@10   & MAP@3      & MAP@5 & MAP@7     & MAP@10 & MRR        \\ \hline
\model & \textbf{80.0\%} &  \textbf{80.0\%}   & \textbf{100.0\%} & \textbf{100.0\%} & \textbf{100.0\%} & \textbf{80.0\%}  & \textbf{84.0\%} & \textbf{88.57\%} & \textbf{92.0\%} &     \textbf{84.0\%}        \\
GNN & 20.0\% &  40.0\%   & 40.0\% & 40.0\% & 60.0\% & 26.67\%  & 32.0\% & 34.29\% & 38.0\% &     30.65\%        \\
PC & 0\% &  20.0\%   & 20.0\% & 40.0\% & 40.0\% & 13.33\%  & 16.0\% & 22.86\% & 28.0\% &     14.0\%        \\
C-LSTM & 0\% &  20.0\%   & 20.0\% & 20.0\% & 20.0\% & 13.33\%  & 16.0\% & 17.14\% & 18.0\% &     10.82\%        \\
Dynotears & 20.0\% &  40.0\%   & 40.0\% & 40.0\% & 40.0\% & 33.33\%  & 36.0\% & 37.14\% & 38.0\% &    30.79\%        \\
GOLEM & 20.0\% &  40.0\%   & 40.0\% & 40.0\% & 40.0\% & 33.33\%  & 36.0\% & 37.14\% & 38.0\% &    31.22\%        \\
\hline
	\end{tabular}
	\label{tab:overall_aiops}
\end{table*}

\subsection{Experimental Setup}
\subsubsection{Datasets}
We evaluated \model\ on the following three real-world datasets for the task of root cause localization.
 \textbf{1) AIOps}: 
This dataset was collected from a real micro-service system.
This system has 234 microservice pods/applications (low-level system entities)  that are deployed to 5 cloud servers (high-level system entities).
The operators collected metrics data (\textit{e.g.}, CPU Usage, Memory Usage) of high-level and low-level system entities from May 2021 to December 2021.
There are 5 system faults during this time period.
\textbf{2)
WADI~\cite{ahmed2017wadi}}: 
This dataset was collected from a water distribution testbed, which owns $3$ stages (high-level entities) and $123$ sensors (low-level entities). 
It has $15$ system faults collected in $16$ days.
In these datasets, low-level entities affiliate with high-level entities, and same-level entities invoke each other.
\textbf{3) 
Swat~\cite{mathur2016swat}}: 
This dataset was collected from a water treatment testbed, which consists of $6$ stages (high-level entities) that have $51$ sensors (low-level entities).
It has $16$ system faults collected in $11$ days.

\subsubsection{Evaluation Metrics}
We evaluated the model performance with the following three widely-used metrics~\cite{meng2020localizing,liu2021microhecl}:

\noindent\textbf{Precision@K (PR@K)}.
It denotes the probability that the top-$K$ predicted root causes are real, defined as
\begin{equation}
    \text{PR@K} = \frac{1}{\mathbb{|A|}}\sum_{a\in \mathbb{A}}\frac{\sum_{i<K}R_a(i)\in V_a}{min(K,|V_a|)},
\end{equation}
where $\mathbb{A}$ is the set of system faults; $a$ is one fault in $\mathbb{A}$; $V_a$ is the real root causes of $a$; $R_a$ is the predicted root causes of $a$; and $i$ refers to the $i$-th predicted cause of $R_a$. 

\noindent\textbf{Mean Average Precision@K (MAP@K)}.
It assesses the model performance in the top-$K$ predicted causes from the overall perspective, defined as
\begin{equation}
    \text{MAP@K} = \frac{1}{K|\mathbb{A}|} \sum_{a\in \mathbb{A}} \sum_{1 \leq j \leq K} \text{PR@j},
\end{equation}
where a higher value indicates better performance. 

\noindent\textbf{Mean Reciprocal Rank (MRR)}.
This metric measures the ranking capability of models. 
The larger the MRR value is, the further ahead the predicted positions of the root causes are; thus, operators can find the real root causes more easily. 
MRR is defined as
\begin{equation}
    \text{MRR} = \frac{1}{\mathbb{A}} \sum_{a\in \mathbb{A}} \frac{1}{rank_{R_a}},
\end{equation}
where $rank_{R_a}$ is the rank number of the first correctly predicted root cause for system fault $a$.

\subsubsection{Baselines}
We compared \model\ with the following five causal discovery models:
\textbf{1) PC}\cite{spirtes2000causation} is a classic constraint-based method.
    It first identifies the skeleton of the causal graph with the independence test, then generates the orientation direction using the v-structure and acyclicity constraints.
\textbf{2) C-LSTM}\cite{tank2021neural} captures the nonlinear Granger causality that existed in multivariate time series by using LSTM neural networks.
\textbf{3) Dynotears}\cite{pamfil2020dynotears} is a score-based method that uses the structural vector autoregression model to construct dynamic Bayesian networks.
 \textbf{4) GOLEM}\cite{ng2020role} employs a likelihood-based score function to relax the hard DAG constraint in NOTEARS. 
\textbf{5) GNN} is a simplified version of our causal discovery method. It only uses GNN to learn causal structures among low-level system entities.


\begin{table*}[!htbp]
\caption{The influence of network propagation in terms of MAP@10}
\begin{tabular}{|c|cc|cc|cc|cc|cc|}
\hline
\multirow{2}{*}{} & \multicolumn{2}{c|}{PC}                          & \multicolumn{2}{c|}{GLOEM}                       & \multicolumn{2}{c|}{Dynotears}                   & \multicolumn{2}{c|}{C-LSTM}                      & \multicolumn{2}{c|}{GNN}                         \\ \cline{2-11} 
                  & \multicolumn{1}{c|}{Original} & Propagate        & \multicolumn{1}{c|}{Original} & Propagate        & \multicolumn{1}{c|}{Original} & Propagate        & \multicolumn{1}{c|}{Original} & Propagate        & \multicolumn{1}{c|}{Original} & Propagate        \\ \hline
Swat              & \multicolumn{1}{c|}{35.0\%}   & \textbf{37.39\%} & \multicolumn{1}{c|}{25.0\%}   & \textbf{33.44\%} & \multicolumn{1}{c|}{30.83\%}  & \textbf{37.08\%} & \multicolumn{1}{c|}{31.87\%}  & \textbf{34.16\%} & \multicolumn{1}{c|}{41.87\%}  & \textbf{49.16\%} \\ \hline
WADI              & \multicolumn{1}{c|}{34.57\%}  & \textbf{35.71\%} & \multicolumn{1}{c|}{34.83\%}  & \textbf{38.05\%} & \multicolumn{1}{c|}{26.81\%}  & \textbf{33.76\%} & \multicolumn{1}{c|}{32.72\%}  & \textbf{42.61\%} & \multicolumn{1}{c|}{37.53\%}  & \textbf{45.98\%} \\ \hline
AIOPS             & \multicolumn{1}{c|}{28.0\%}   & \textbf{30.0\%}  & \multicolumn{1}{c|}{38.0\%}   & \textbf{54.0\%}  & \multicolumn{1}{c|}{38.0\%}   & \textbf{58.0\%}  & \multicolumn{1}{c|}{18.0\%}   & \textbf{48.0\%}  & \multicolumn{1}{c|}{38.0\%}   & \textbf{60.0\%}  \\ \hline
\end{tabular}
\label{np_rca_map10}
\end{table*}

\begin{table*}[!htbp]
\caption{The influence of network propagation in terms of MRR}
\begin{tabular}{|c|cc|cc|cc|cc|cc|}
\hline
\multirow{2}{*}{} & \multicolumn{2}{c|}{PC}                          & \multicolumn{2}{c|}{GLOEM}                       & \multicolumn{2}{c|}{Dynotears}                   & \multicolumn{2}{c|}{C-LSTM}                      & \multicolumn{2}{c|}{GNN}                         \\ \cline{2-11} 
                  & \multicolumn{1}{c|}{Original} & Propagate        & \multicolumn{1}{c|}{Original} & Propagate        & \multicolumn{1}{c|}{Original} & Propagate        & \multicolumn{1}{c|}{Original} & Propagate        & \multicolumn{1}{c|}{Original} & Propagate        \\ \hline
Swat              & \multicolumn{1}{c|}{26.16\%}  & \textbf{32.27\%} & \multicolumn{1}{c|}{22.36\%}  & \textbf{30.42\%} & \multicolumn{1}{c|}{27.85\%}  & \textbf{33.98\%} & \multicolumn{1}{c|}{29.35\%}  & \textbf{32.85\%} & \multicolumn{1}{c|}{34.77\%}  & \textbf{40.43\%} \\ \hline
WADI              & \multicolumn{1}{c|}{27.74\%}  & \textbf{30.74\%} & \multicolumn{1}{c|}{23.48\%}  & \textbf{25.89\%} & \multicolumn{1}{c|}{22.22\%}  & \textbf{34.28\%} & \multicolumn{1}{c|}{24.39\%}  & \textbf{33.27\%} & \multicolumn{1}{c|}{32.71\%}  & \textbf{36.40\%} \\ \hline
AIOPS             & \multicolumn{1}{c|}{14.0\%}   & \textbf{25.35\%} & \multicolumn{1}{c|}{31.22\%}  & \textbf{37.74\%} & \multicolumn{1}{c|}{30.79\%}  & \textbf{50.77\%} & \multicolumn{1}{c|}{10.82\%}  & \textbf{24.73\%} & \multicolumn{1}{c|}{30.65\%}  & \textbf{62.48\%} \\ \hline
\end{tabular}
\label{np_rca_mrr}
\end{table*}

Since none of the above baselines can be directly applied to learn the hierarchical interdependent causation, we only utilized the entity metrics to construct causation between low-level system entities and the system KPI.
We then selected the top-$K$ entities with the highest causal scores as the root causes. 
To verify the effectiveness of the network propagation module (see Section~\ref{sec_network_prop}), we applied it to the causal structures learned by these baselines and analyzed  model performance changes.

In addition, to study the impact of each technical component of \model, we developed the following model variants:
(1) To assess the benefits of inter-level learning (see Section~\ref{topological_rca}), we implemented \model-N by removing the inter-level learning in topological causal discovery while keeping the intra-level learning of low-level system entities, network propagation and individual causal discovery.
(2) To evaluate the necessity and effectiveness of integrating the individual and topological causal discovery, we developed two variants: \model-I, which only keeps the individual causal discovery, and \model-T, which solely keeps the topological causal discovery.
(3) To verify the efficacy of hierarchical GNN-based causal discovery, we replaced the causal discovery component of \model\ with PC, C-LSTM, Dynotears, and GOLEM, respectively, to implement model variants denoted as \model-P, \model-C, \model-D, and \model-G.

All experiments were conducted on a server running Ubuntu 18.04.5 with Intel(R) Xeon(R) Silver 4110 CPU @ 2.10GHz, 4-way GeForce RTX 2080 Ti GPUs, and 192 GB memory. In addition, all methods were implemented using Python 3.8.12 and PyTorch 1.7.1.
\subsection{Performance Evaluation}
\subsubsection{Overall Performance}
\label{overall_sec}
Table~\ref{tab:overall_swat}, Table~\ref{tab:overall_wadi}, and Table~\ref{tab:overall_aiops} present the overall performance of all models, where a larger value indicates better performance.
We have two key observations: 
First, \model\ can significantly outperform all the baselines on all three datasets.
For example, compared to the second-best method, \model\ can improve PR@10, MAP@10, and MRR by at least 21.9\%, 15.7\%, and 6.29\%, respectively.
The underlying driver is that \model\ can capture more complex malfunctioning effects by integrating individual and topological analyses, and learning interdependent causal networks.
Second, GNN is the best baseline model that outperforms others on most datasets.
A possible explanation is that graph neural networks can facilitate the learning of non-linear causal relations among system entities via message passing.
Thus, the experimental results on three datasets demonstrate the superiority of \model\ in locating root causes over other  baselines.

\subsubsection{Influence of Network Propagation}
Here, we applied our network propagation mechanism (see Section~\ref{sec_network_prop}) to the causal structures learned by each baseline model to evaluate its effect on performance.
The results are shown in Table~\ref{np_rca_map10} and Table~\ref{np_rca_mrr}.
Our first finding is that network propagation can always improve the model performance for all models on all datasets.
This observation strongly supports our assumption that network propagation is beneficial for capturing the propagation patterns of malfunctioning effects, resulting in a superior root cause localization performance.
Moreover, we observe that across all models, network propagation yields greater performance enhancement on AIOps than the other two datasets.
A possible reason is that AIOps contains  explicit invoking relations among different pods, resulting in learning stronger causation compared with Swat and WADI.  

\begin{figure}[!t]
\centering
\subfigure[Precision@K for Swat]{
\includegraphics[width=4.1cm]{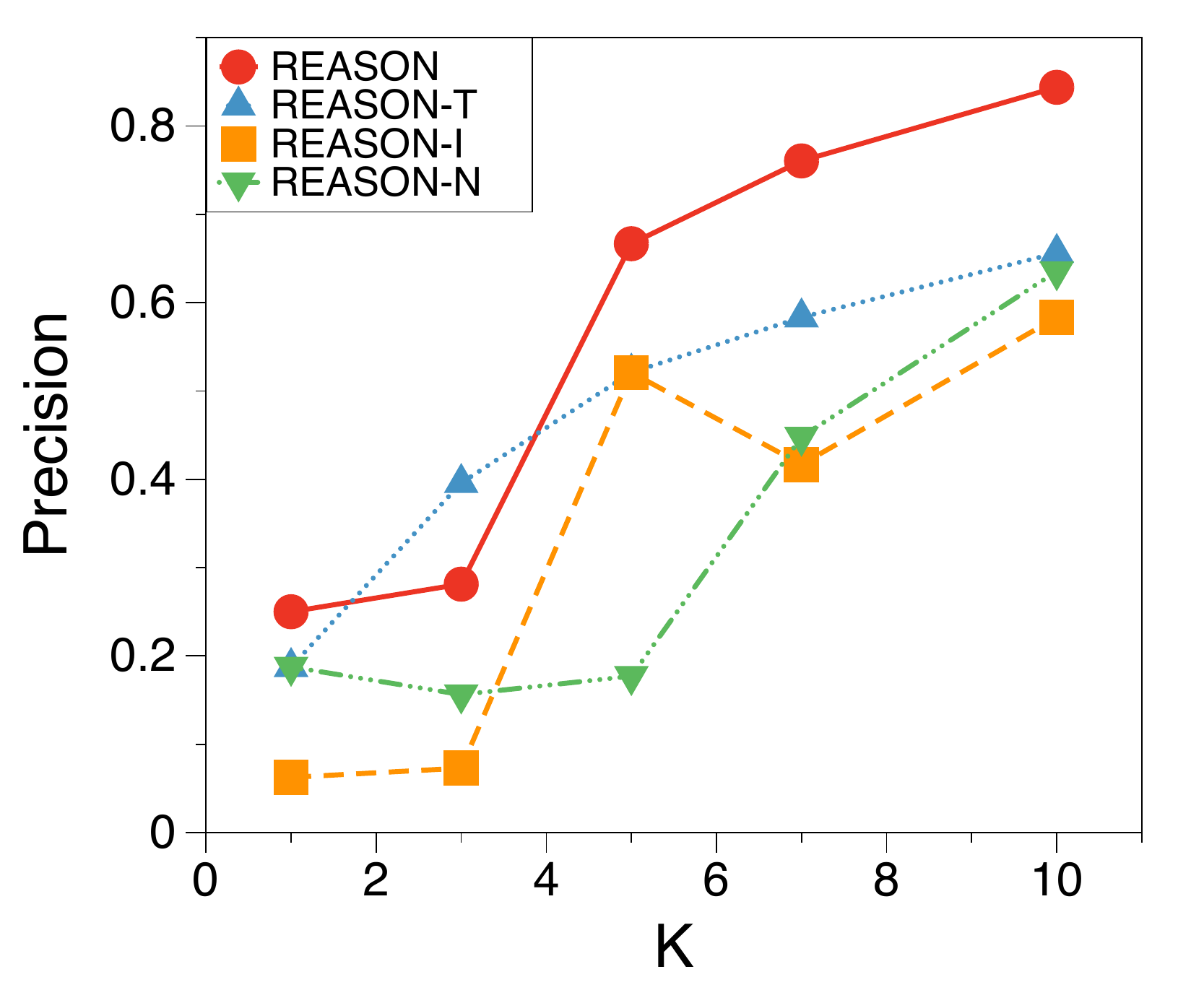}
}
\hspace{-2mm}
\subfigure[MAP@K for Swat]{ 
\includegraphics[width=4.1cm]{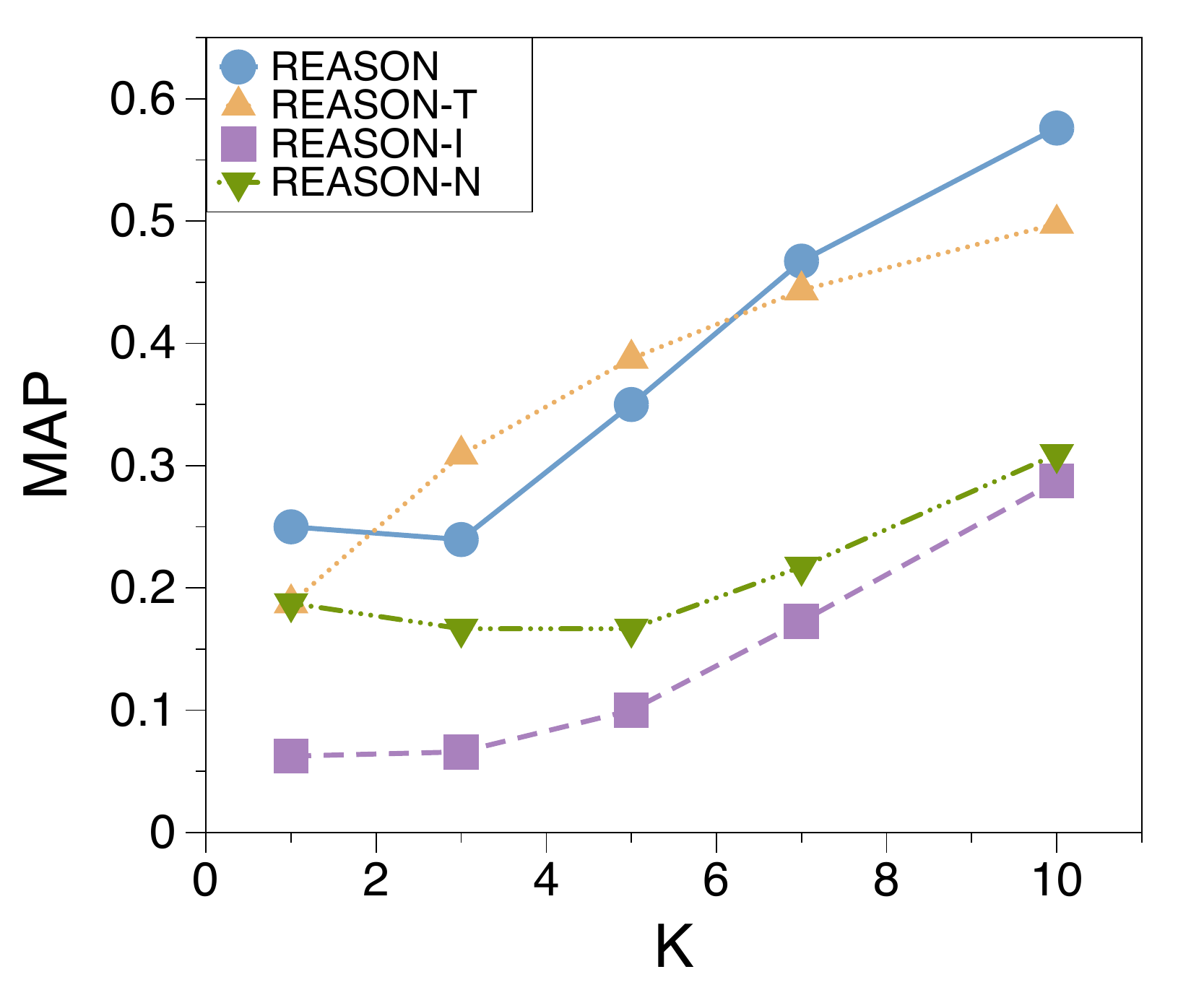}
}

\subfigure[Precision@K for AIOps]{
\includegraphics[width=4.1cm]{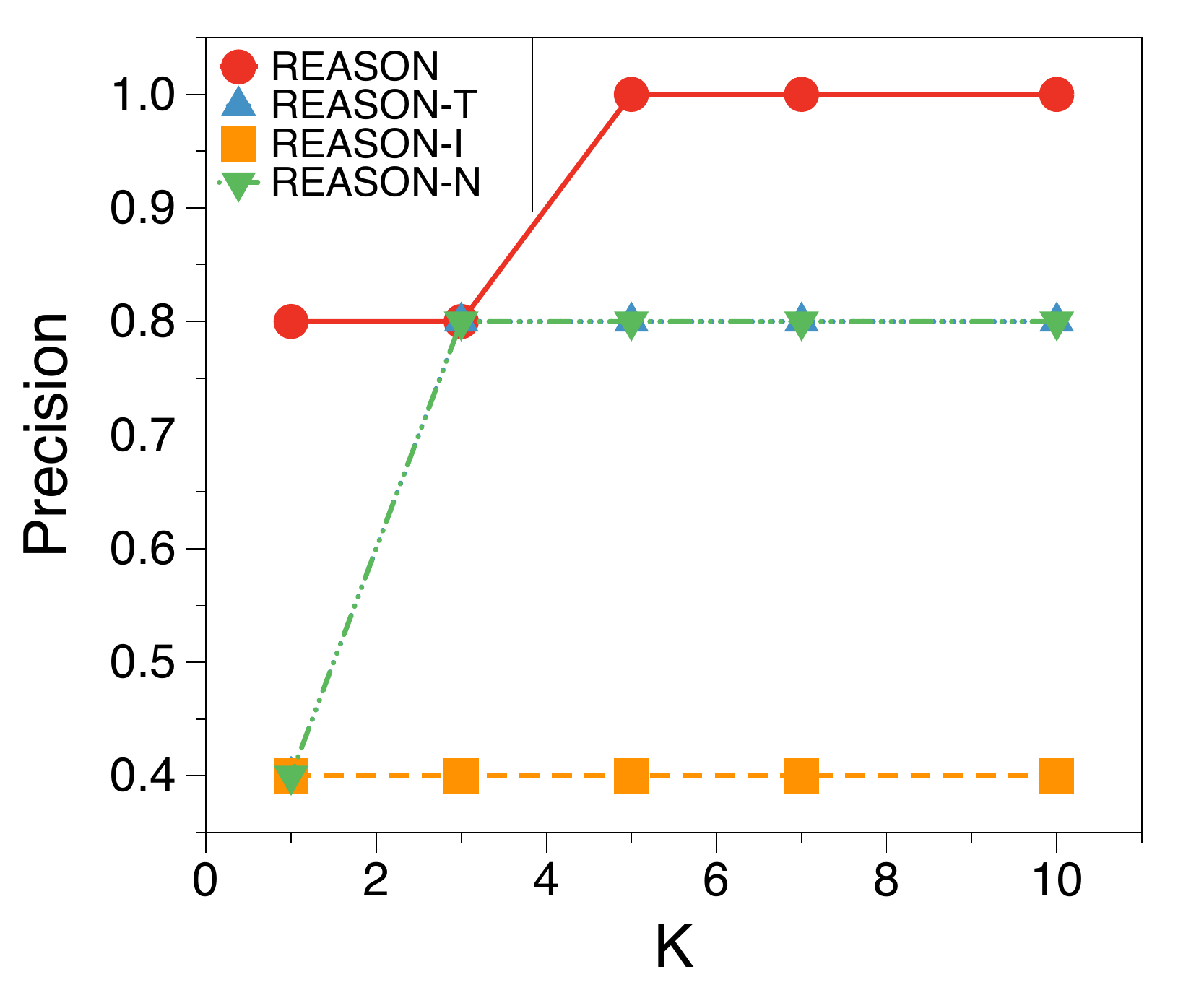}
}
\hspace{-2mm}
\subfigure[Precision@K for WADI]{
\includegraphics[width=4.1cm]{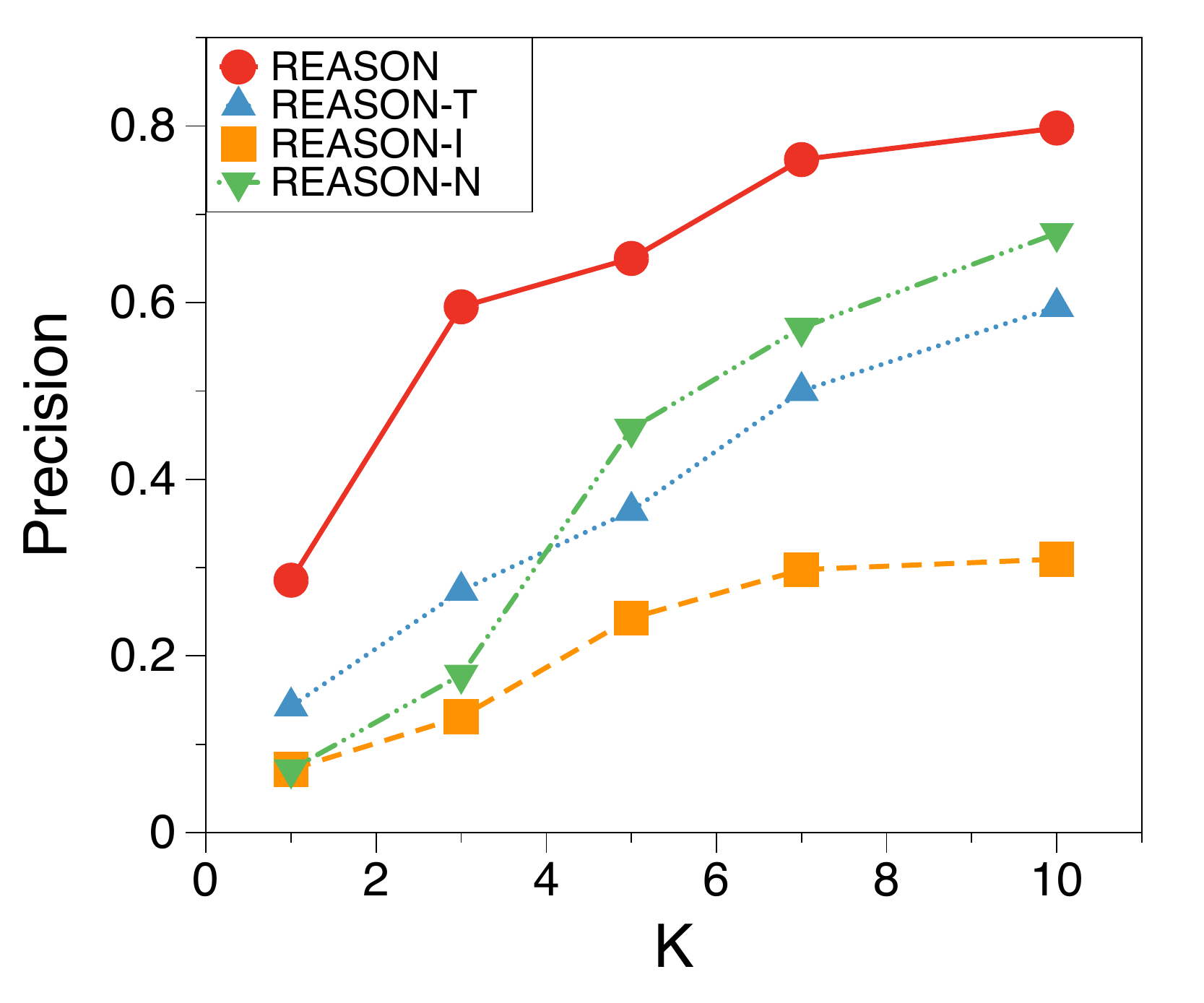}
}

\subfigure[MAP@K for AIOps]{ 
\includegraphics[width=4.1cm]{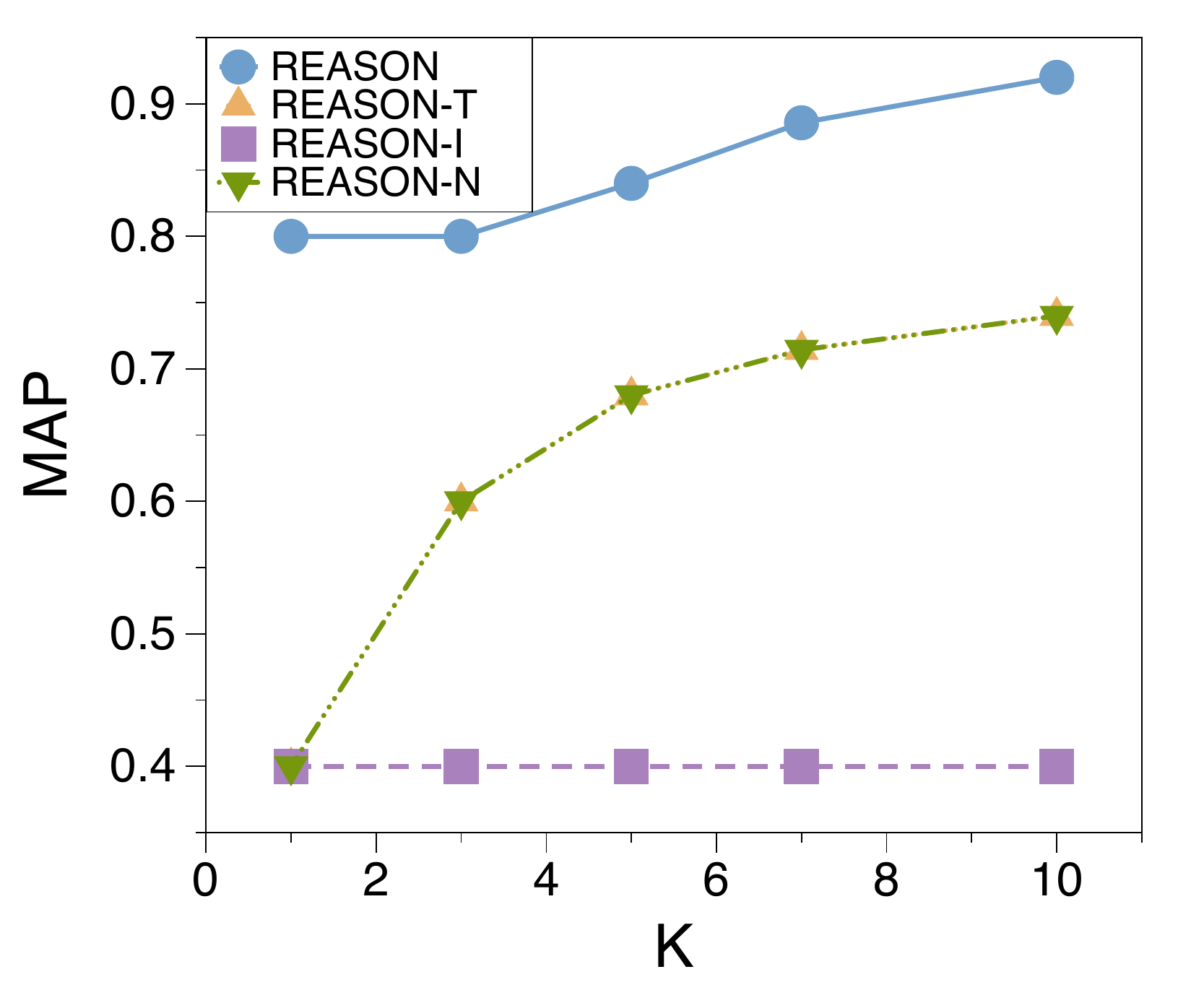}
}
\hspace{-2mm}
\subfigure[MAP@K for WADI]{ 
\includegraphics[width=4.1cm]{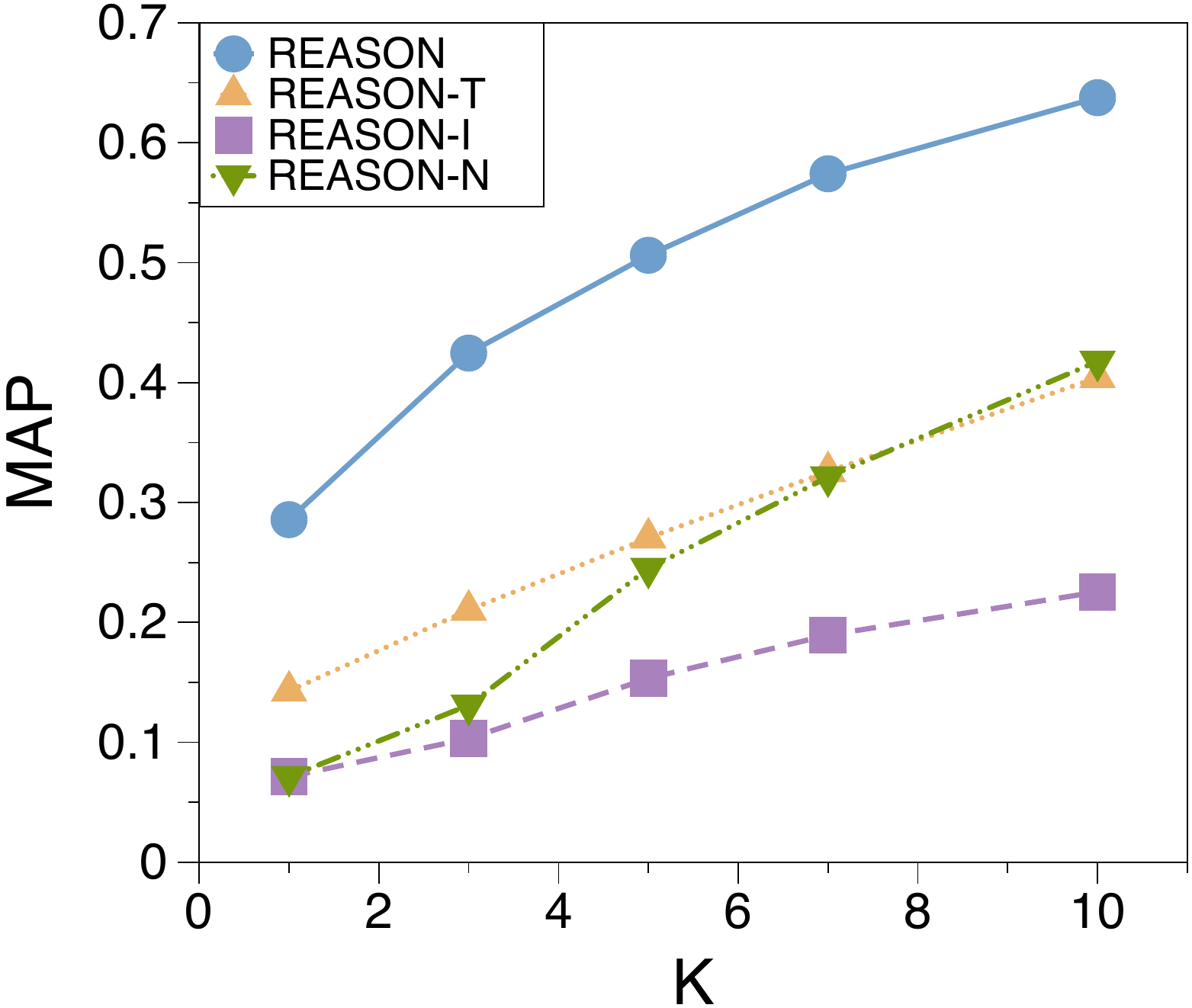}
}
\caption{
Ablation studies of \model.}
\label{fig:it_rca}
\end{figure}

\subsubsection{Ablation studies of \model}
\label{subsec:necessityICDTCD}
Figure~\ref{fig:it_rca} shows ablation studies of \model\ to examine the necessity of each technical component using PR@K and MAP@K.
We can find that \model\  significantly outperforms \model-N on Swat and WADI.
The underlying driver is that since \model-N focuses on modeling causation among low-level entities only, using such causal structures, \model-N is unable to capture cross-network propagation patterns of malfunctioning effects, leading to worse model performance.
The second finding is that \model\ is superior to both \model-T and \model-I in most cases.
This observation indicates that integrating individual and topological causal discovery results can sufficiently capture the fluctuation and propagation patterns of malfunctioning effects for precisely locating root causes.
Thus, each technical component of \model\ is indispensable for keeping excellent root cause localization performance.

\begin{figure}[!t]
 \vspace{-10pt}
\centering
\subfigure[Precision@K for Swat]{
\includegraphics[width=4.1cm]{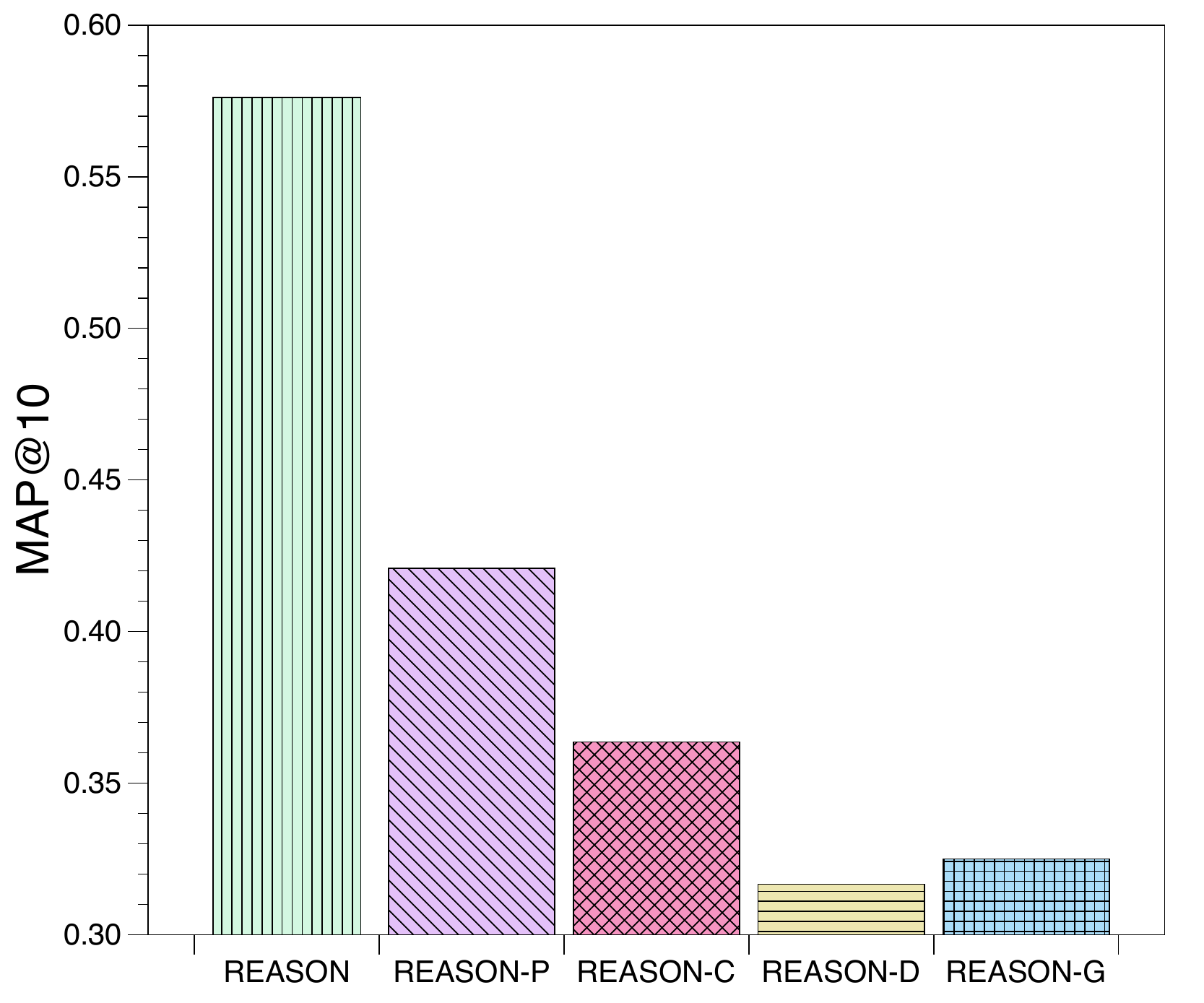}
}
\hspace{-2mm}
\subfigure[MAP@K for Swat]{ 
\includegraphics[width=4.1cm]{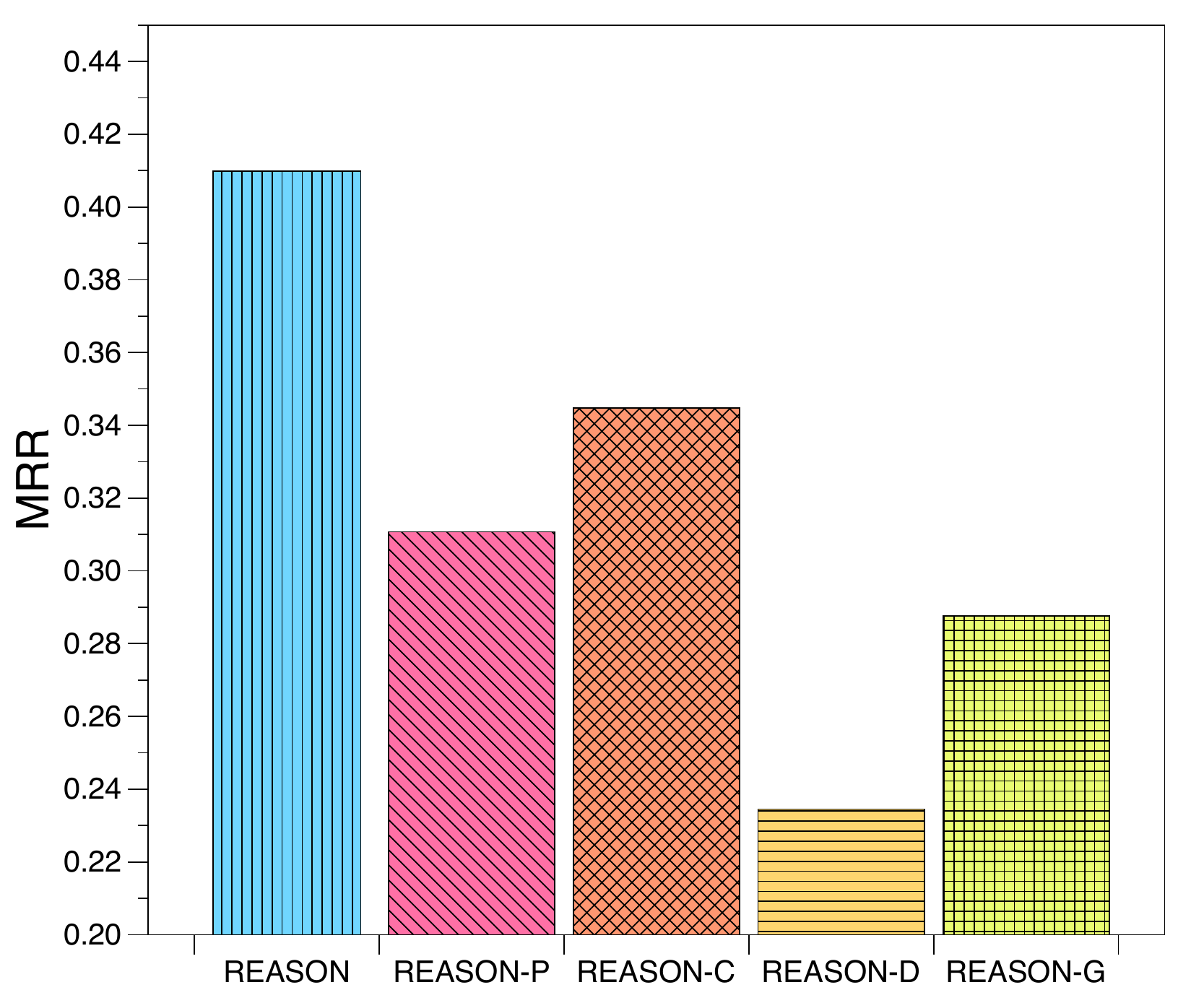}
}\vspace{-3pt}

\subfigure[MAP@10 for AIOps]{
\includegraphics[width=4.1cm]{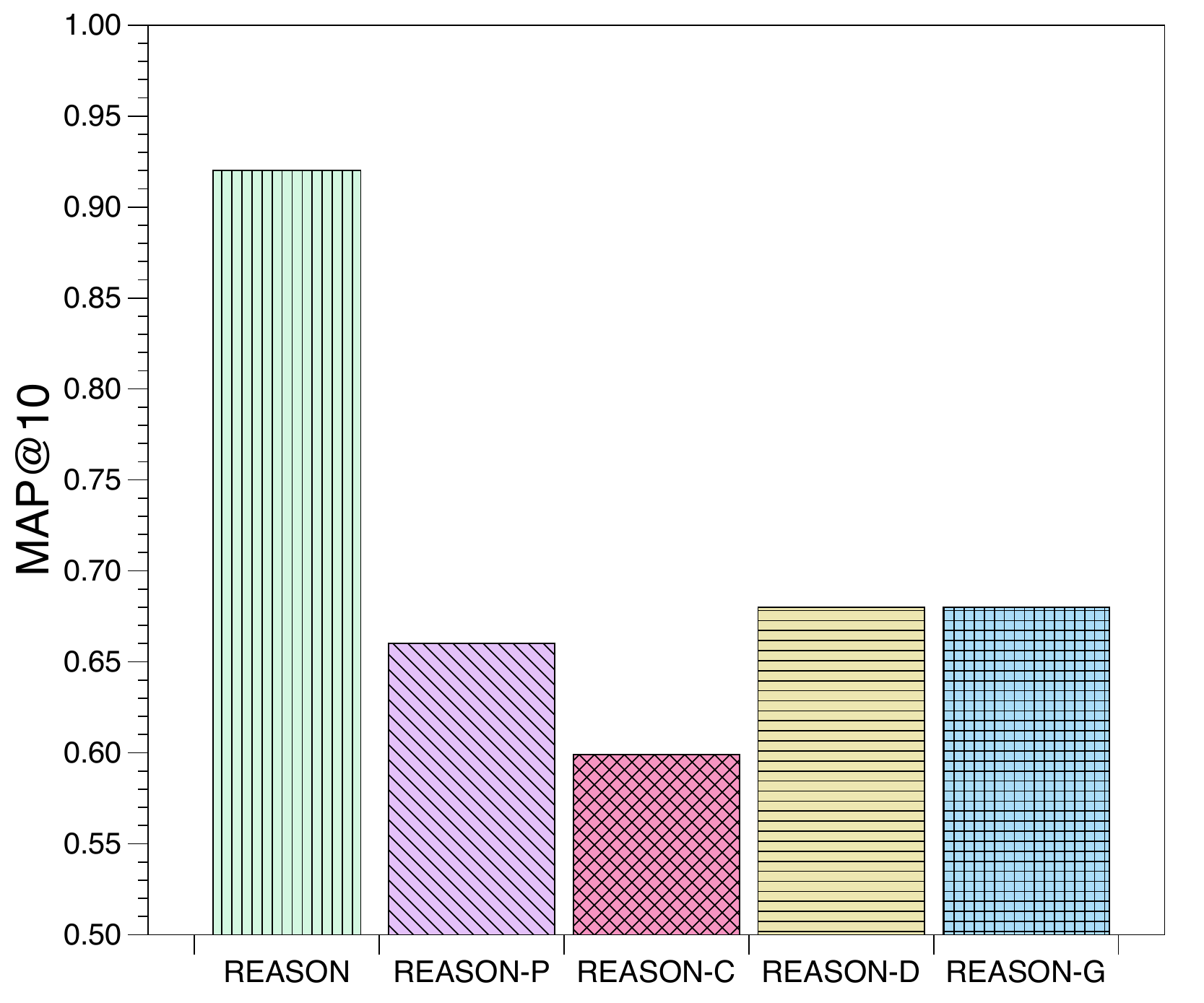}
}
\hspace{-2mm}
\subfigure[MRR for AIOps]{ 
\includegraphics[width=4.1cm]{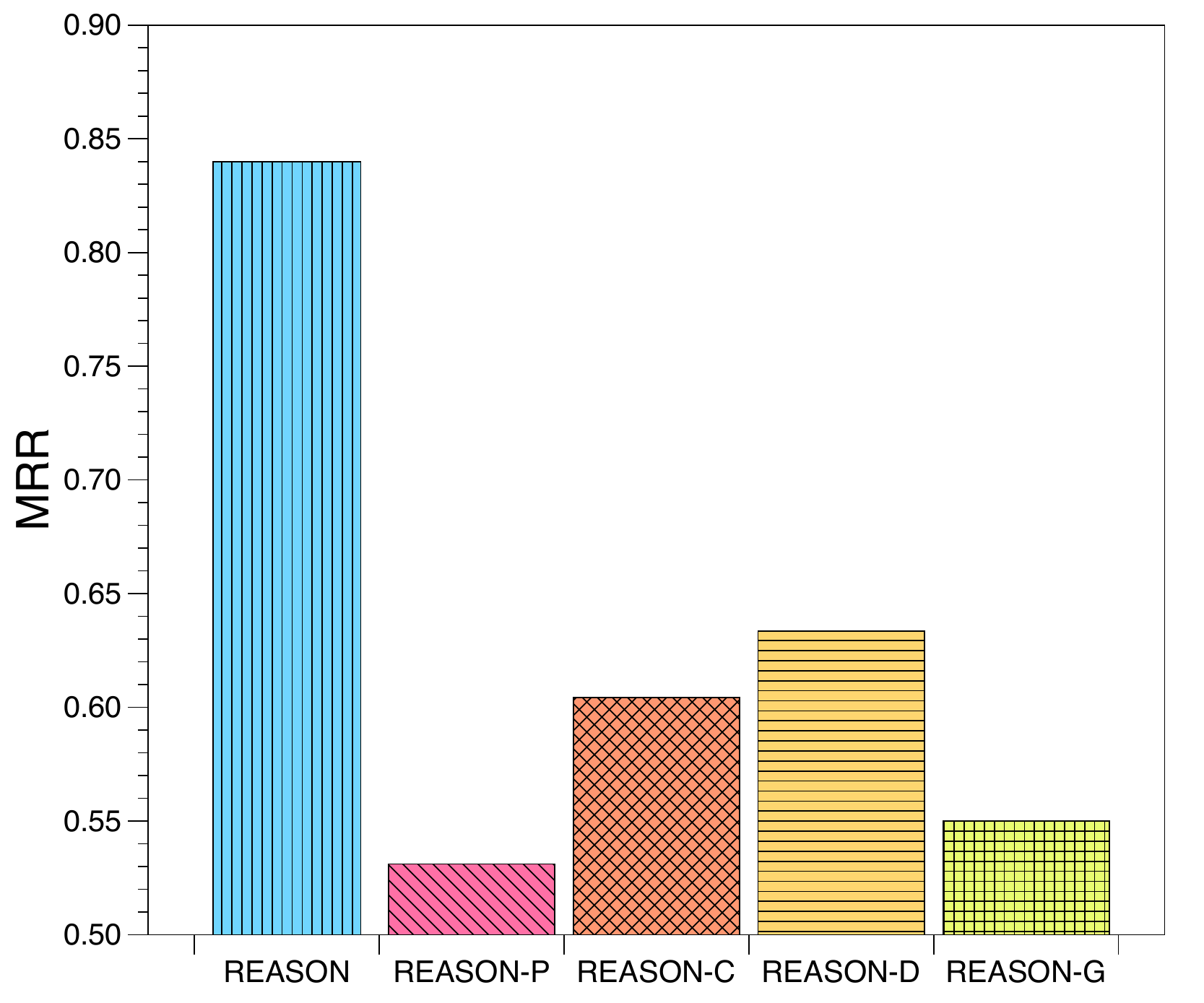}
}\vspace{-3pt}

\subfigure[MAP@10 for WADI]{
\includegraphics[width=4.1cm]{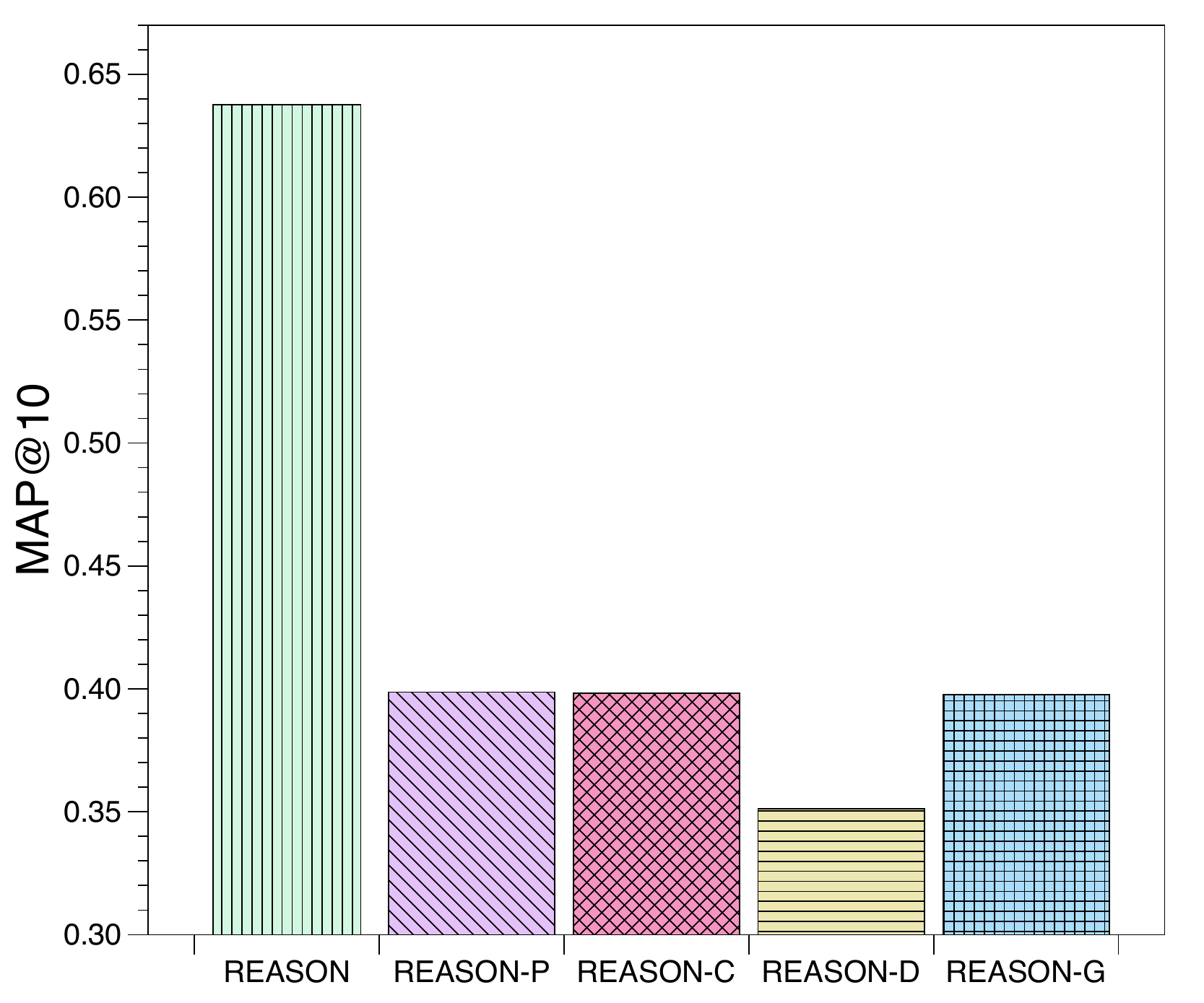}
}
\hspace{-2mm}
\subfigure[MRR for WADI]{ 
\includegraphics[width=4.1cm]{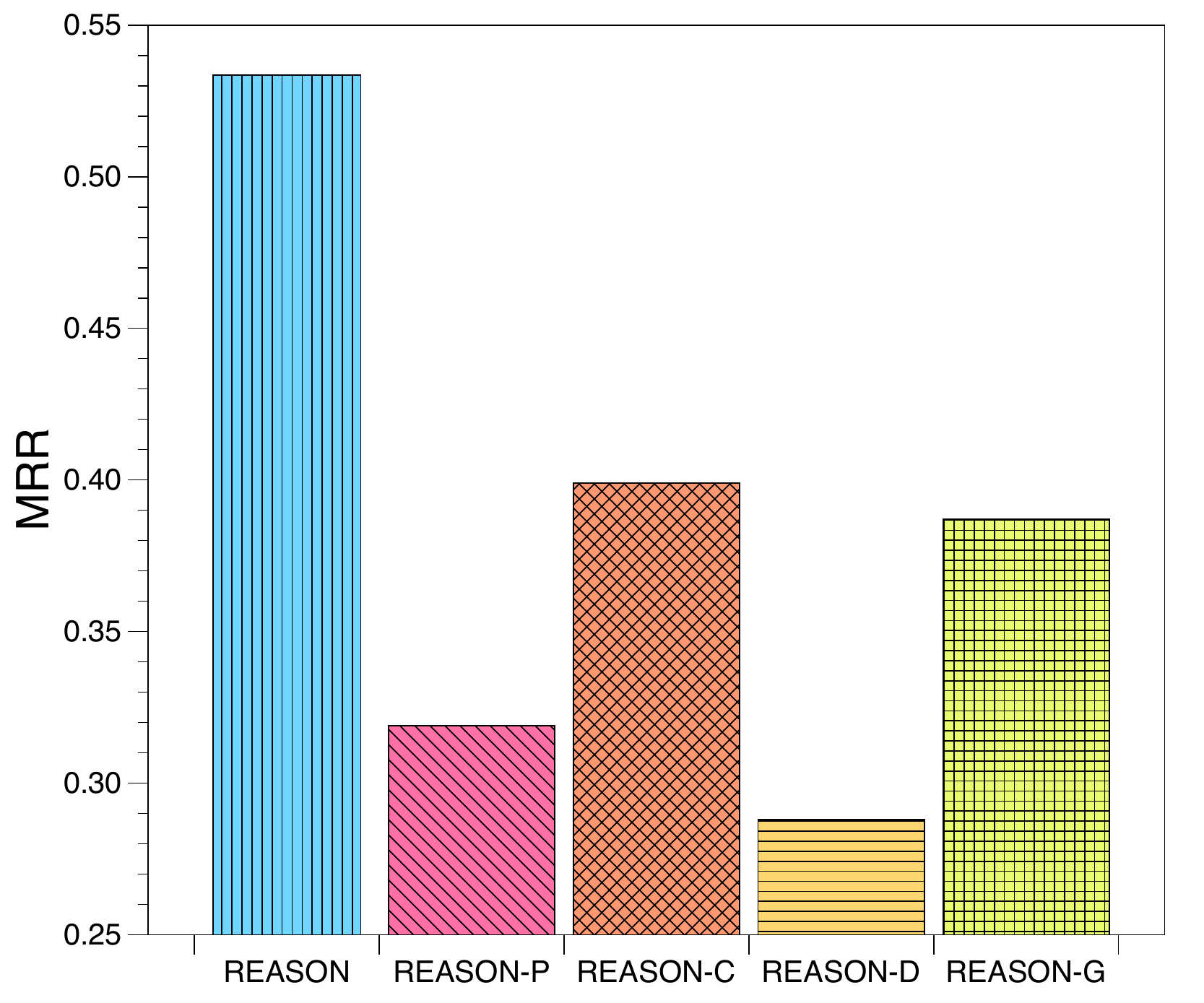}
}
\vspace{-10pt}
\caption{The impact of hierarchical GNN-based causal learning process.}
\label{fig:cd_rca}
\vspace{-5pt}
\end{figure}

\subsubsection{Impact of Hierarchical GNN-based Causal Discovery}
Figure~\ref{fig:cd_rca} evaluates the effectiveness of the proposed hierarchical GNN-based causal discovery method.
Our key observations are two-fold.
First, we find that \model\ significantly outperforms all model variants.
The underlying driver is that the message-passing mechanism of GNN can learn more robust non-linear causal relations through sharing neighborhood information.
Moreover, \model-P outperforms \model-C across all datasets in terms of MAP@10, while the result is the opposite in terms of MRR. 
A possible explanation is that PC learns more causal relations between system entities than C-LSTM.
As a result, \model-P is able to identify more actual root causes by propagating on the learned causal structures, but their ranks are not at the top owing to more root cause candidates.

\begin{figure}[t]
 \vspace{-10pt}
\centering
\subfigure[$\gamma$ for Swat]{
\includegraphics[width=4.1cm]{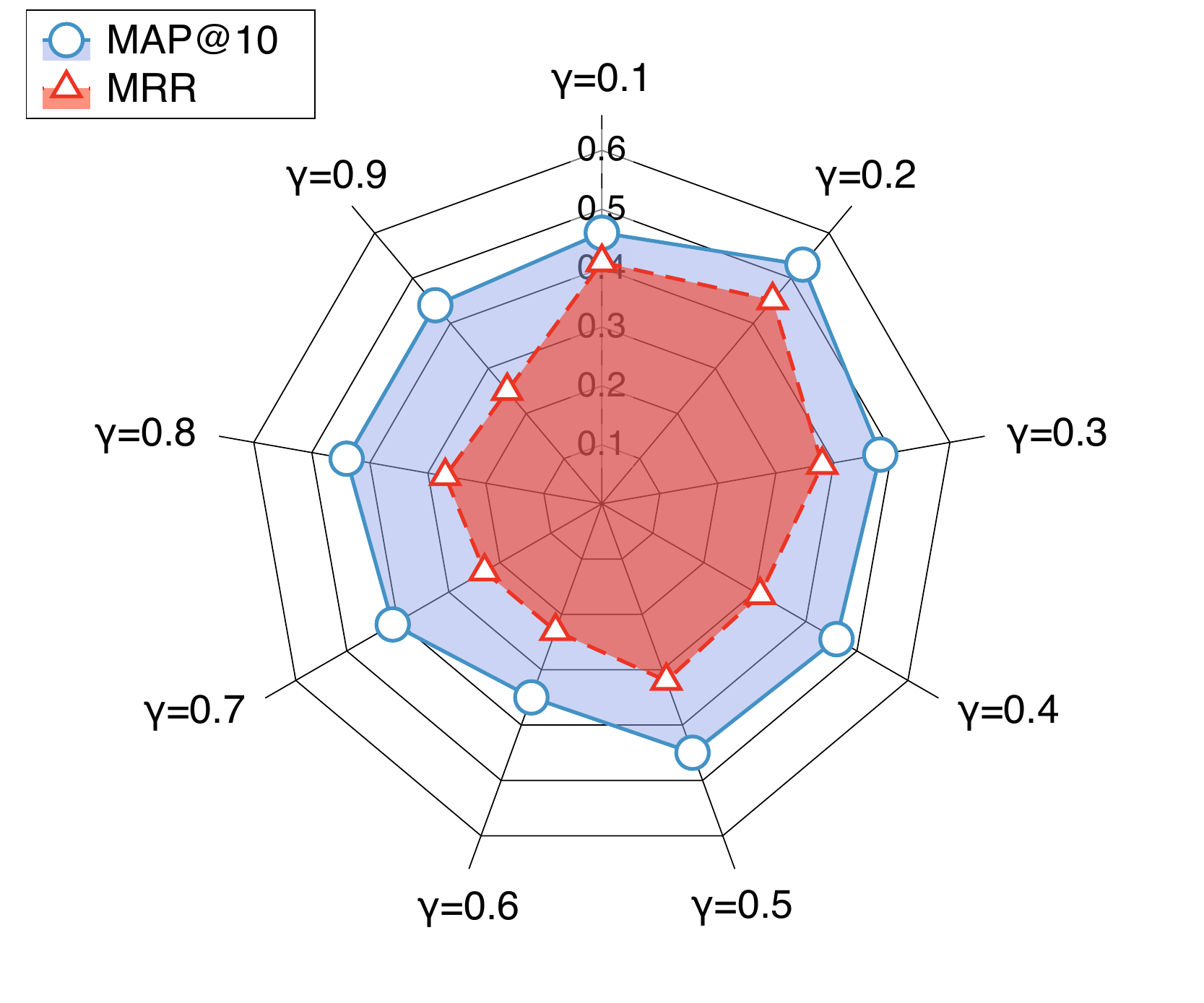}
}
\hspace{-3mm}
\subfigure[$L$ for Swat]{
\includegraphics[width=4.1cm]{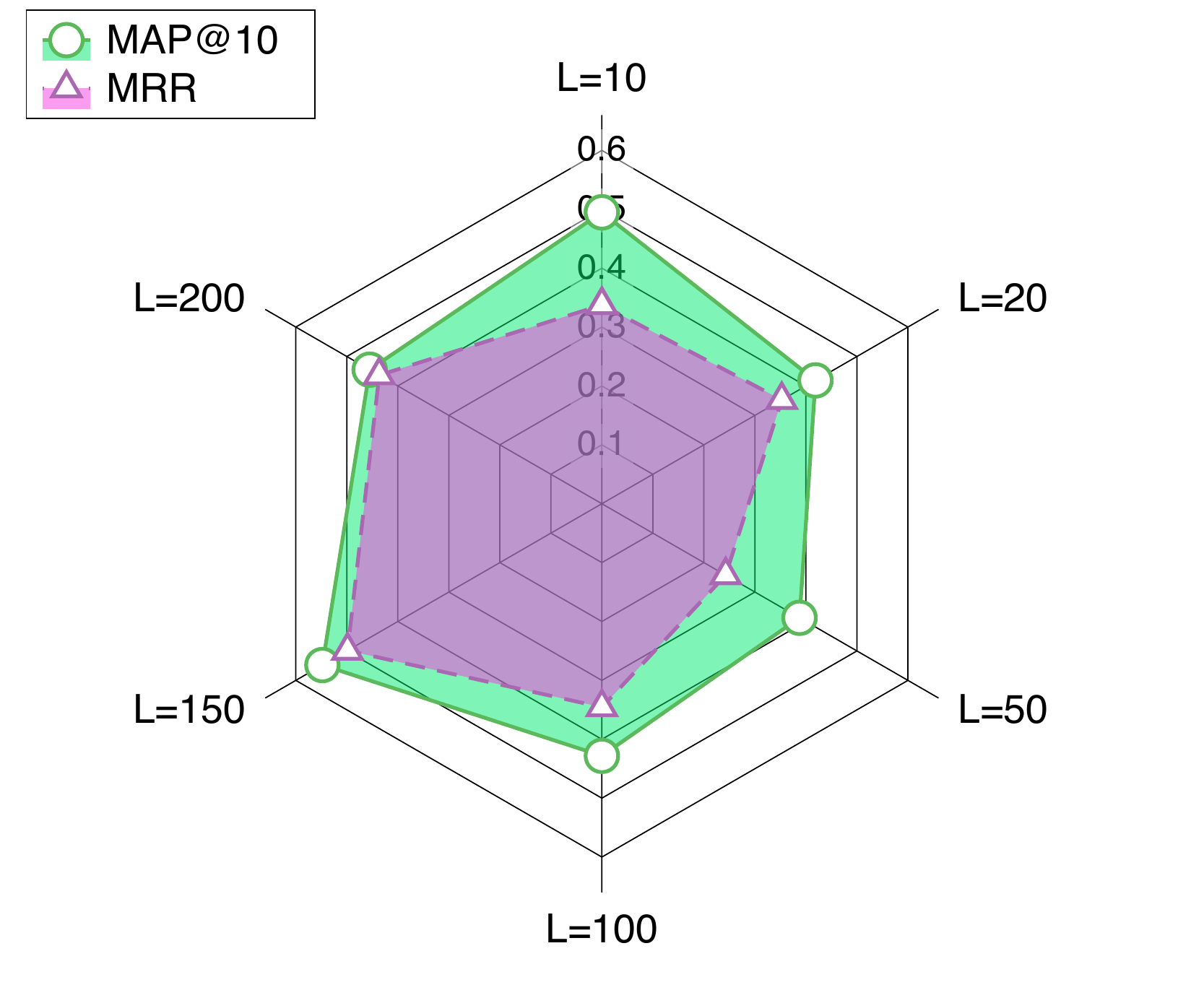}
}\vspace{-3pt}

\subfigure[$\gamma$ for WADI]{ 
\includegraphics[width=4.1cm]{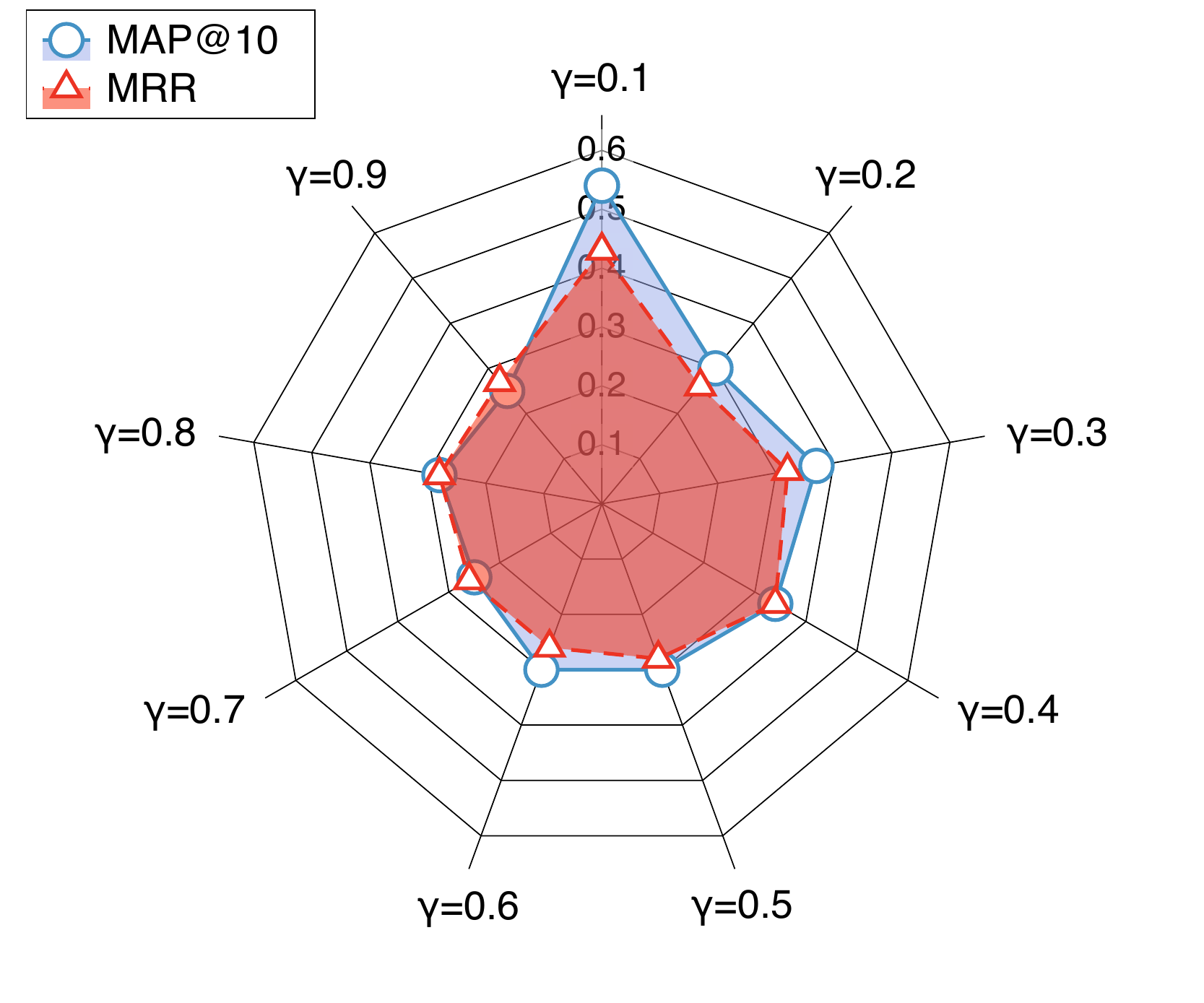}
}
\hspace{-3mm}
\subfigure[$L$ for WADI]{ 
\includegraphics[width=4.1cm]{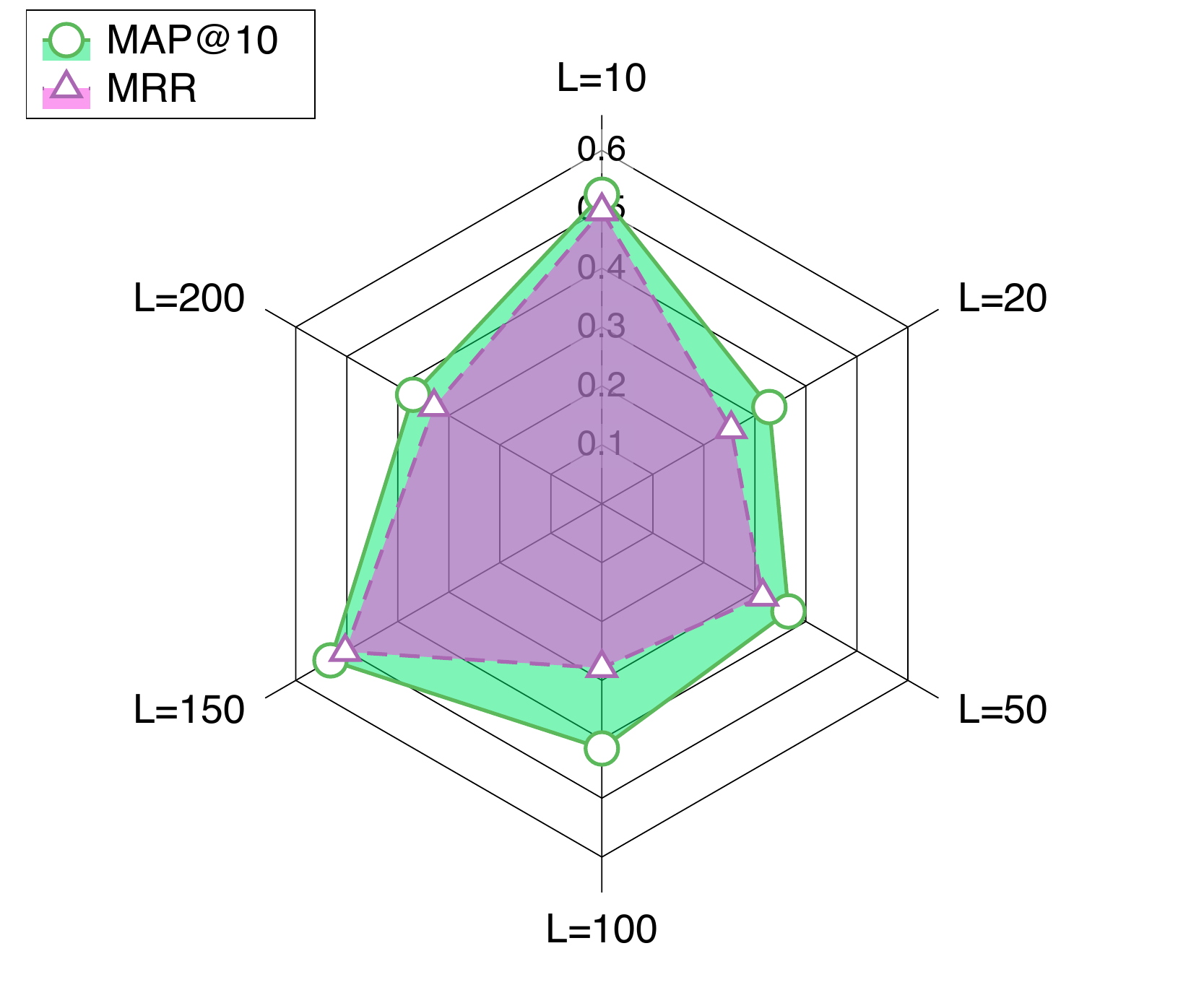}
}\vspace{-3pt}

\subfigure[$\gamma$ for AIOps]{ 
\includegraphics[width=4.1cm]{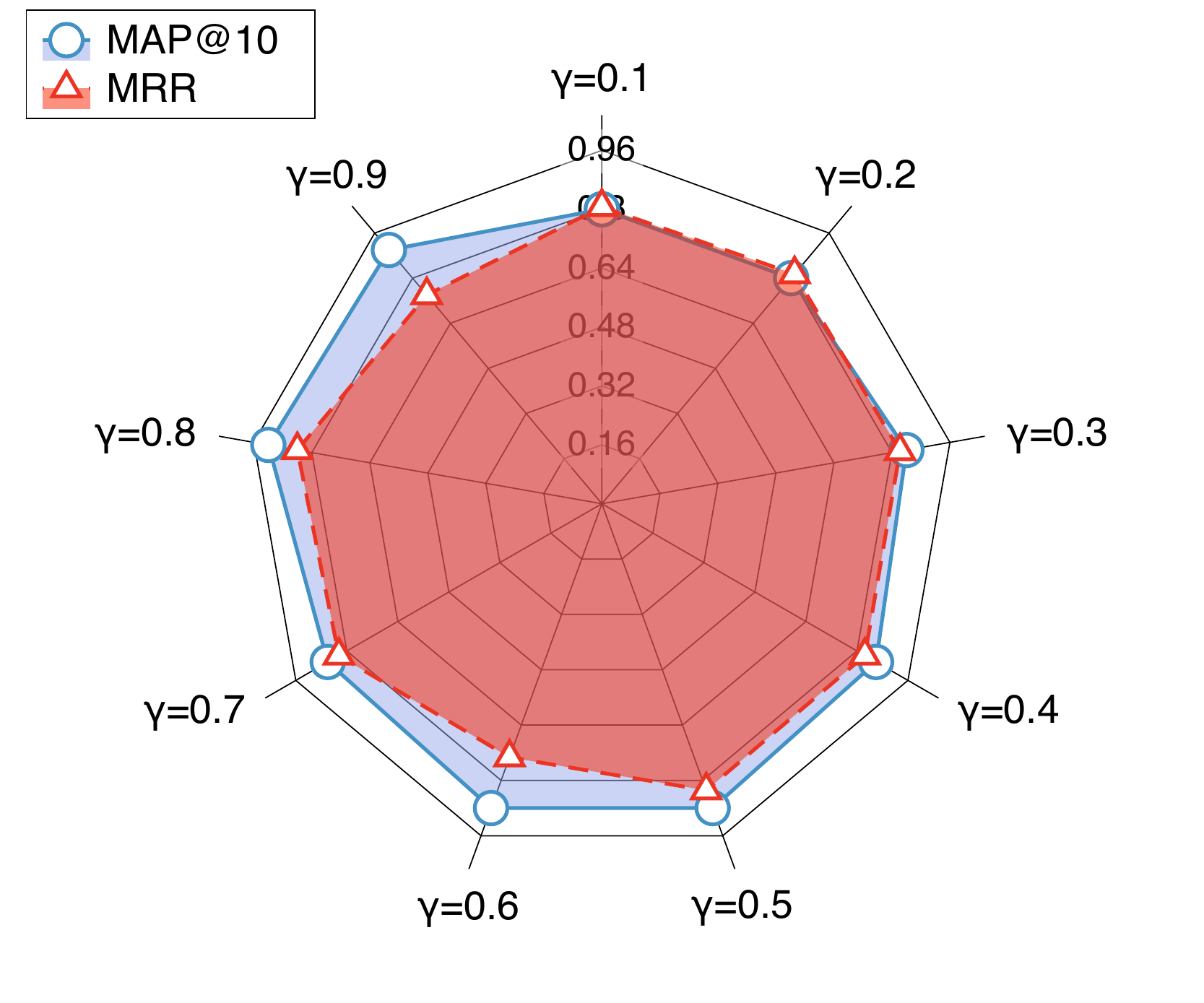}
}
\hspace{-3mm}
\subfigure[$L$ for AIOps]{ 
\includegraphics[width=4.1cm]{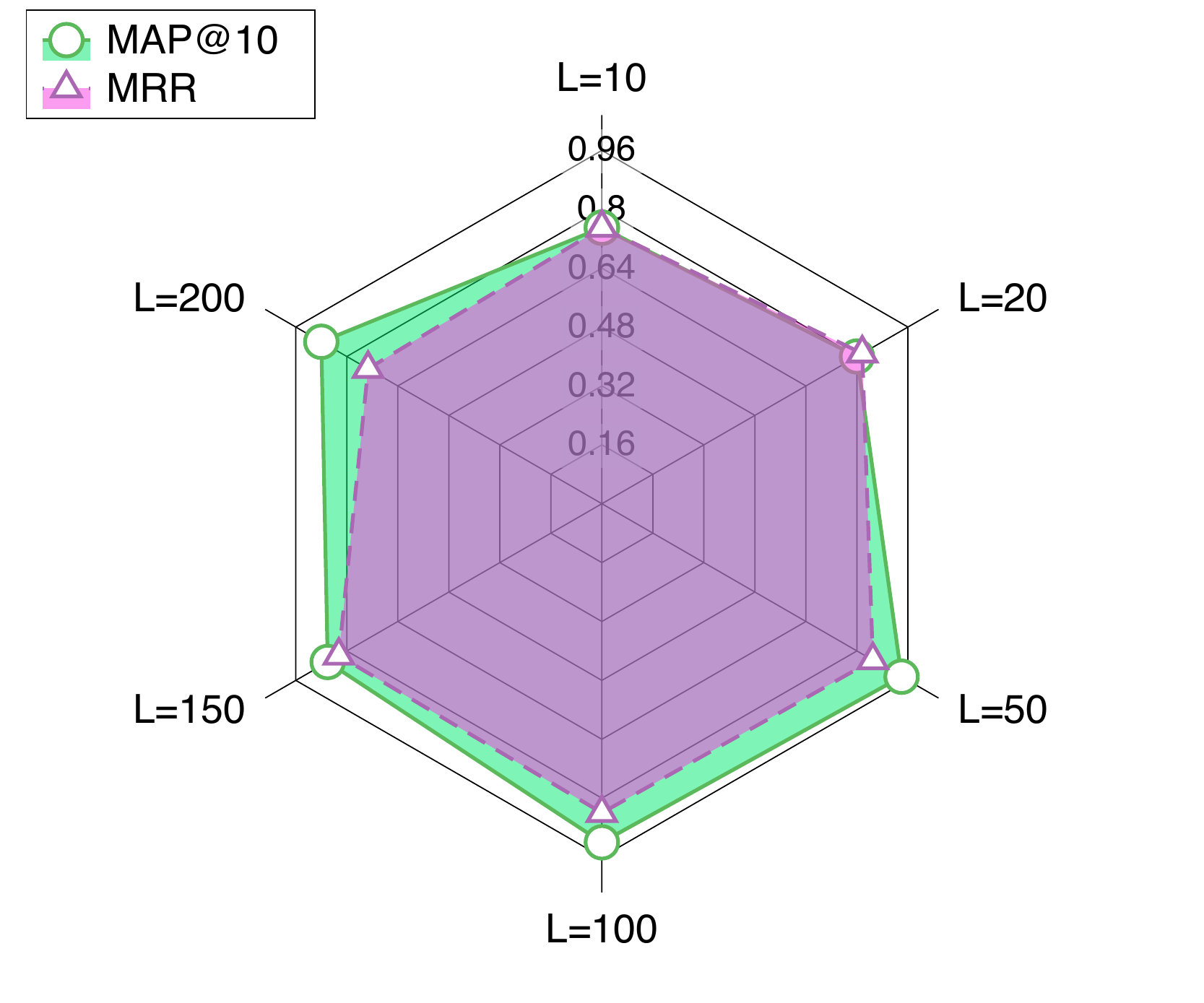}
}
 \vspace{-10pt}
\caption{Parameter analysis of \model.}
\label{fig:parameters}
\vspace{-5pt}
\end{figure}

\subsubsection{Parameter Analysis.}
\label{pa_appendix}
We investigated the integration parameter $\gamma$ and the number of layers $L$ in GNN.
$\gamma$ controls the contribution of the individual and topological causal discovery for root cause localization.
The number of layers $L$ in GNN impacts the learning scenario of interdependent causal structures.
Figure~\ref{fig:parameters} presents our parameter analysis results. 
It can be seen that although the optimal $\gamma$ values for different datasets vary, \model\ can achieve optimal or near-optimal results on all three datasets using a similar small $\gamma$ value.    
For instance, the best value of $\gamma$ for Swat, WADI, and AIOps is 0.2, 0.1, and 0.8, respectively. But when we use $\gamma = 0.1$, the best value for WADI, the overall performance of \model\ only drops a little in terms of both MAP@10 and MRR. For instance, compared with the optimal results, the MRR value only decreased by 0.01 on Swat and 0.04 on AIOps, respectively. This indicates that although the propagation of malfunctioning effects varies amongst datasets, the topological component contributes more to the model performance than the individual component, which further supports our findings in Section~\ref{subsec:necessityICDTCD}. Thus, in most cases, a small $\gamma$ value (\textit{e.g.}, $\gamma = 0.1$) would be a good choice. Second, when the number of GNN layers rose, we did not observe improved model performance. This is because a large number of GNNs may cause the information of each node to become highly similar, hindering the learning of robust causal relationships.

\subsubsection{A Case Study}
Finally, we conducted a case study to further illustrate the learned interdependent causal networks by utilizing the system failure of AIOps on September 1, 2021.
Operators built up a microservice system and simulated system faults to collect metrics data for analysis.
The detailed collection procedure is as follows: 
First, the operators deployed the system on three servers that are \textit{compute-2}, \textit{infra-1}, and \textit{control-plane-1}.
Then, they sent requests periodically to the pod \textit{sdn-c7kqg} to observe the system's latency.
Next, to simulate the malfunctioning effects of the root cause, the operators used an \textit{opennssl} command to make the pod \textit{catalogue-xfjp} have an extremely high CPU load, which affected some other pods on different servers, and eventually caused the system fault.
Finally, the operators collected all entity metrics (\textit{e.g.}, CPU Usage, Memory Usage) of all system entities (\textit{e.g.}, servers, pods).

\begin{figure}[!t]
    \centering
    \includegraphics[width=0.7\linewidth]{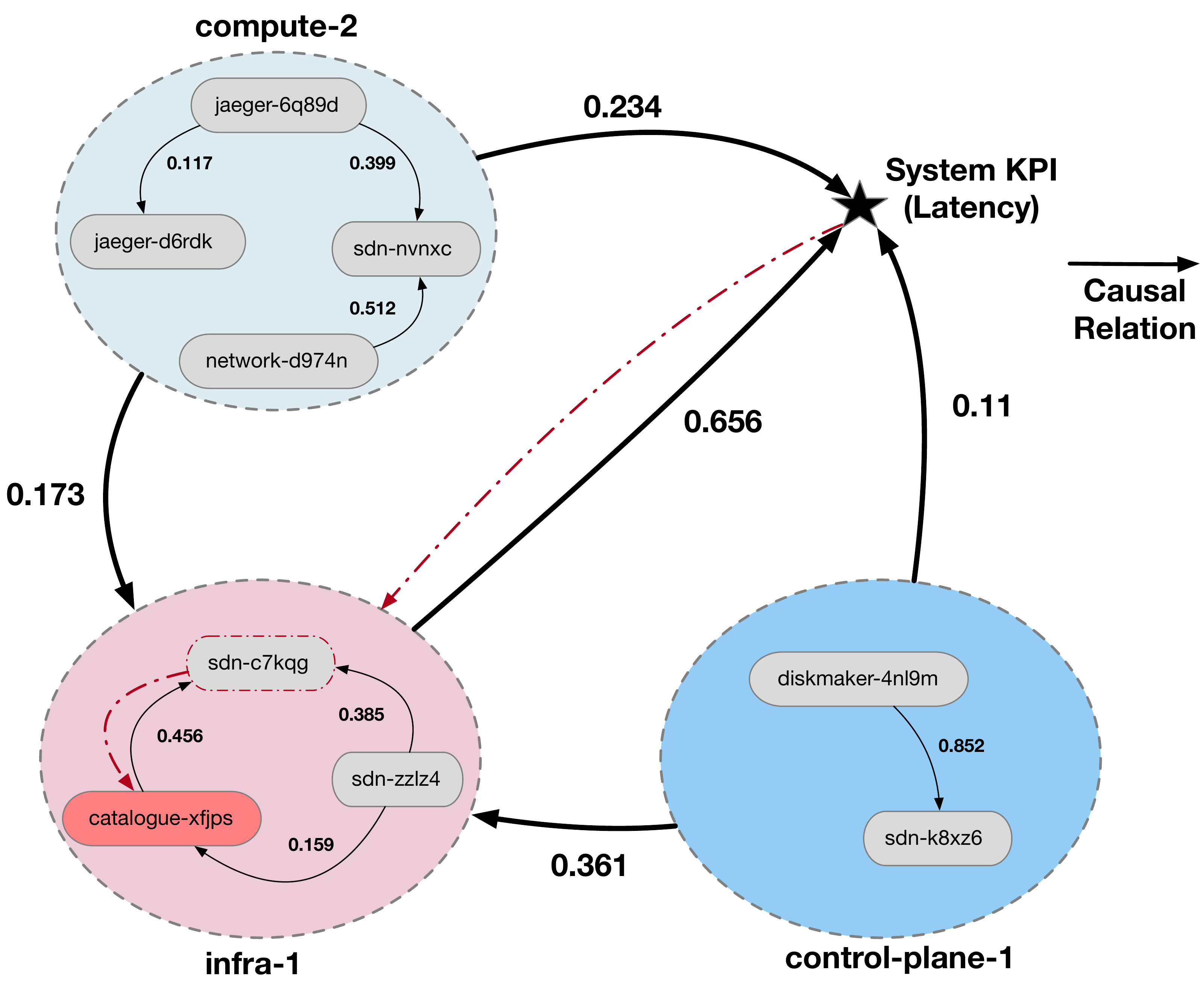}
    \captionsetup{justification=justified, singlelinecheck=off}
    \caption{Interdependent causal structures learned from AIOps dataset. Applications/pods are denoted by solid nodes, in which the red solid node is the root cause. The numbers on each causal relation edge with black color indicates the causal score between two connected nodes. The red dashed line reflects the backtracing process of the root cause.}
    \label{fig:causal_graph}
\end{figure}

Based on the collected metrics data, we applied \model\ to learn the interdependent causation between system entities and the system KPI for locating root causes, which reflects the real operation circumstances.
Figure~\ref{fig:causal_graph} shows the learned interdependent causal structures based on the CPU Usage metric.
According to it, \textit{infra-1} server is the one most likely to increase in system latency.
In this server, \textit{catalogue-xfjp} is the root cause, whose negative effects propagate to `\textit{sdn-c7kqg}, resulting in the malfunction of \textit{infra-1}.
This observation illustrates that \model\ can precisely locate the root causes and provide an explanation for the located outcomes.

\nop{\section{Related Works}}
\section{Related Work}

\noindent\textbf{Root Cause Analysis (RCA)}, also known as fault localization, focuses on identifying the root causes of system failures/faults\nop{by inferring} from symptom observations~\cite{sole2017survey}.
In recent years, many domain-specific RCA approaches~\cite{fourlas2021survey,soldani2022anomaly,deng2021graph} have been proposed for maintaining the robustness of complex systems in various domains.
For instance, in the energy management domain, Capozzoli {\it{et al.}} utilized statistical techniques and DNNs to identify the reason for abnormal energy consumption in smart buildings~\cite{capozzoli2015fault}.
In the web development domain, Brandon {\it et al.} proposed a graph representation framework to localize root causes in microservice systems by comparing anomalous situations and graphs~\cite{brandon2020graph}. Different from the existing works, the proposed \model\ framework is a generic RCA approach that analyzes the surveillance multi-variate time series data from both individual and topological perspectives. Moreover, \model\ captures the interdependent network properties present in many real-world systems to enhance RCA performance.

\noindent\textbf{Causal Discovery in Time Series}  aims to learn causal relationships from observational time series data~\cite{assaad2022survey}. 
Existing methods can be broadly classified into four categories: (i) Granger causality approaches~\cite{nauta2019causal,tank2021neural}, in which the causation is assessed based on whether one time series is helpful in predicting another;
(ii) Constraint-based approaches~\cite{runge2020discovering,sun2015causal,entner2010causal}, in which a causal structure is learned based on the  conditional independence test and v-structure rules;
(iii) Noise-based approaches~\cite{hyvarinen2010estimation,peters2013causal}, in which the causation is depicted by equations that reflect the causation between different variables and noises; 
(iv) Score-based approaches~\cite{pamfil2020dynotears,bellot2021neural}, in which a causal structure's quality is assessed by a scoring function. \model\ belongs to the score-based causal discovery category. Existing causal discovery methods could only handle time series or isolated graphs, ignoring the structural and dynamical features one may want to model explicitly. In this paper, we propose a hierarchical graph neural networks based method that could model interdependent causal structures from multi-variable time series.

\noindent\textbf{Interdependent Networks}  are often referred to as network of networks (NoN), in which complex networks interact and influence one another~\cite{HAMICHE2016319,NEKOVEE2007457}. 
Numerous real-world systems exhibit such structural and dynamical features that differ from those observed in isolated networks. To overcome the limitation of prior efforts on isolated graph analysis, in recent years, increasing research efforts have been focused on interdependent networks and their applications. For example, Ni {\it{et al.}} employed interdependent networks to illustrate the academic influence of scholars based on their research area and publications~\cite{ni2014inside}.
Laird {\it{et al.}} studied the interdependent relationship between cancer pain and depression~\cite{laird2009cancer}. 
These examples demonstrate the efficiency of modeling complicated systems via interdependent networks.
In recent years, several studies~\cite{7563819,6956877,online_failure} have begun to explore how the interdependent networks model can be applied to root cause analysis.
However, there are two key differences between \model\ and other previous works:
1) Existing works only consider physical or statistical correlations, but not causation. 
2) Existing interdependent networks are constructed using domain knowledge or system rules.
\model\ can automatically discover the interdependent causal graphs from monitoring metrics data for root cause analysis.

\section{Conclusion}
\nop{We present a generic RCA framework that localizes the root causes of system faults through mining interdependent causal relations between system components and system KPI automatically.
This framework consists of two sub-models: individual causal discovery (ICA) and topological causal discovery (TCA).
The ICA is to assess each component by analyzing the abnormal pattern in its monitoring system metric in isolation.
The TCA is to construct the causal relations of same-level and cross-level system entities to imitate the propagation mode of system faults for evaluating each component.
Through extensive experiments, we can find that our approach is effective at discovering the root causes of system faults and providing interdependent causal graphs among system entities.
Moreover, hierarchical structures existing in real-world systems significantly improve the performance of root cause localization.
In addition, imitating the propagation of anomalous patterns is beneficial to find hidden real root causes.
In the future, to find root causes earlier,  we will bring incremental learning to the RCA domain in order to efficiently and effectively conduct RCA.}

In this paper, we investigated the challenging problem of root cause localization in complex systems with interdependent network structures. We proposed \model, a generic framework for root cause localization through mining interdependent causation and propagation patterns of faulted effects. Hierarchical graph neural networks were used to represent non-linear intra-level and inter-level causation and to improve causal discovery among system entities via message transmission. 
We conducted comprehensive experiments on three real-world datasets to evaluate the proposed framework. The experimental results validate the effectiveness of our work. Additionally, through ablation studies, parameter analysis, and case studies, the importance of capturing interdependent structures for root cause localization has been well verified. An interesting direction for further exploration would be incorporating other sources of data, such as system logs, with the time series data for root cause analysis in complex systems.

\newpage
\bibliographystyle{ACM-Reference-Format}
\bibliography{acmart}

\end{document}